\documentclass[sigconf,authorversion,nonacm]{acmart}

\usepackage{subfigure}
\usepackage{algorithm}
\usepackage{algpseudocode}

\usepackage{multirow}
\usepackage{color, colortbl}

\newcolumntype{P}[1]{>{\centering\arraybackslash}p{#1}}
\newcolumntype{M}[1]{>{\centering\arraybackslash}m{#1}}
\newcolumntype{R}[1]{>{\raggedleft\arraybackslash}p{#1}}

\begin{document}

\title{Adversarial Feature Alignment: Balancing Robustness and Accuracy in Deep Learning via Adversarial Training}

\author{Leo Hyun Park, Jaeuk Kim, Myung Gyo Oh, Jaewoo Park, and Taekyoung Kwon}
\authornote{Corresponding author.}
\affiliation{
  \institution{Graduate School of Information, Yonsei University}
  \city{Seoul}
  \country{South Korea}}
\email{{dofi, freak0wk, myunggyo.oh, jaewoo1218, taekyoung}@yonsei.ac.kr}
\renewcommand{\shortauthors}{Park et al.}

\begin{abstract}
  Deep learning models continue to advance in accuracy, yet they remain vulnerable to adversarial attacks, which often lead to the misclassification of adversarial examples. Adversarial training is used to mitigate this problem by increasing robustness against these attacks. However, this approach typically reduces a model's standard accuracy on clean, non-adversarial samples. 
The necessity for deep learning models to balance both robustness and accuracy for security is obvious, but achieving this balance remains challenging, and the underlying reasons are yet to be clarified. 

This paper proposes a novel adversarial training method called \textit{Adversarial Feature Alignment (AFA)}, to address these problems. Our research unveils an intriguing insight: \textit{misalignment within the feature space often leads to misclassification, regardless of whether the samples are benign or adversarial}. AFA mitigates this risk by employing a novel optimization algorithm based on contrastive learning to alleviate potential feature misalignment. Through our evaluations, we demonstrate the superior performance of AFA. The baseline AFA delivers higher robust accuracy than previous adversarial contrastive learning methods while minimizing the drop in clean accuracy to 1.86\% and 8.91\% on CIFAR10 and CIFAR100, respectively, in comparison to cross-entropy. We also show that joint optimization of AFA and TRADES, accompanied by data augmentation using a recent diffusion model, achieves state-of-the-art accuracy and robustness.
\end{abstract}


\maketitle

\section{Introduction}

\begin{figure*}[t]
  \centering
  \subfigure[Robustness-Accuracy Tradeoff of DNNs]{
    \includegraphics[width = 0.43\textwidth]{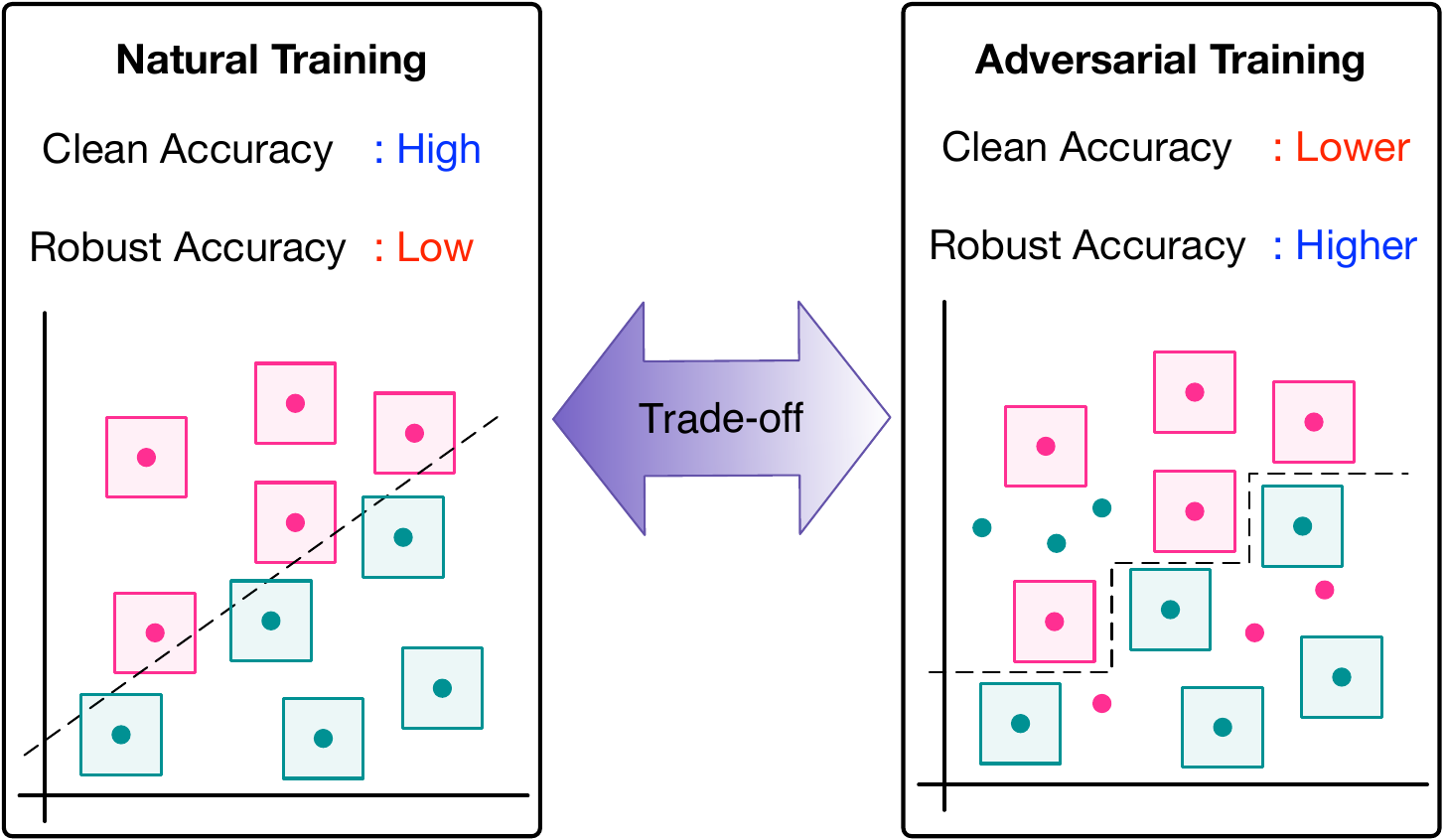}
    \label{fig:introduction:tradeoff}
  }
  \centering
  \subfigure[Misaligned distribution (not enough)]{
    \includegraphics[width = 0.25\textwidth]{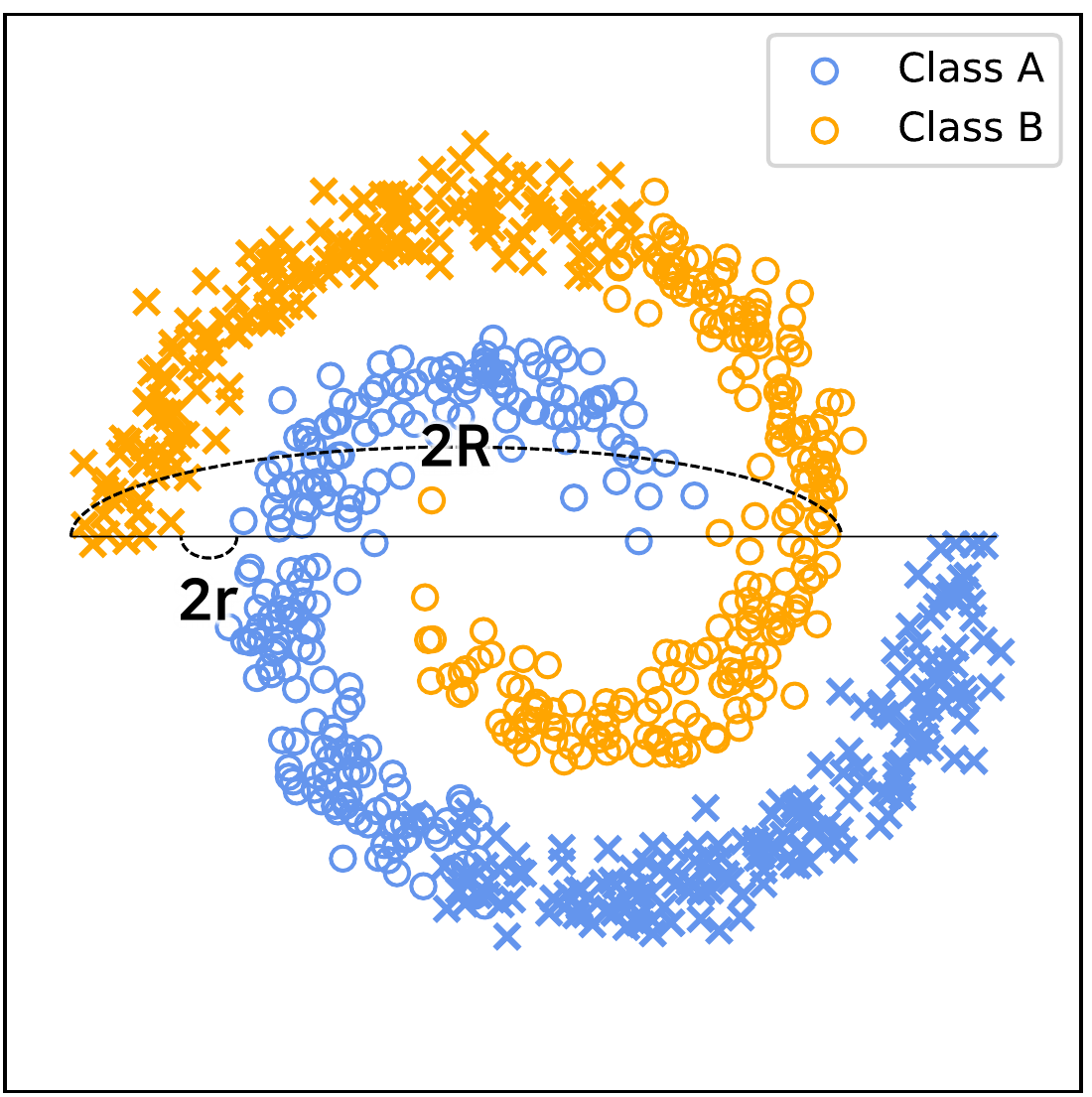}
    \label{fig:introduction:separation}
  }
  \subfigure[Aligned distribution (helpful)]{
    \includegraphics[width = 0.25\textwidth]{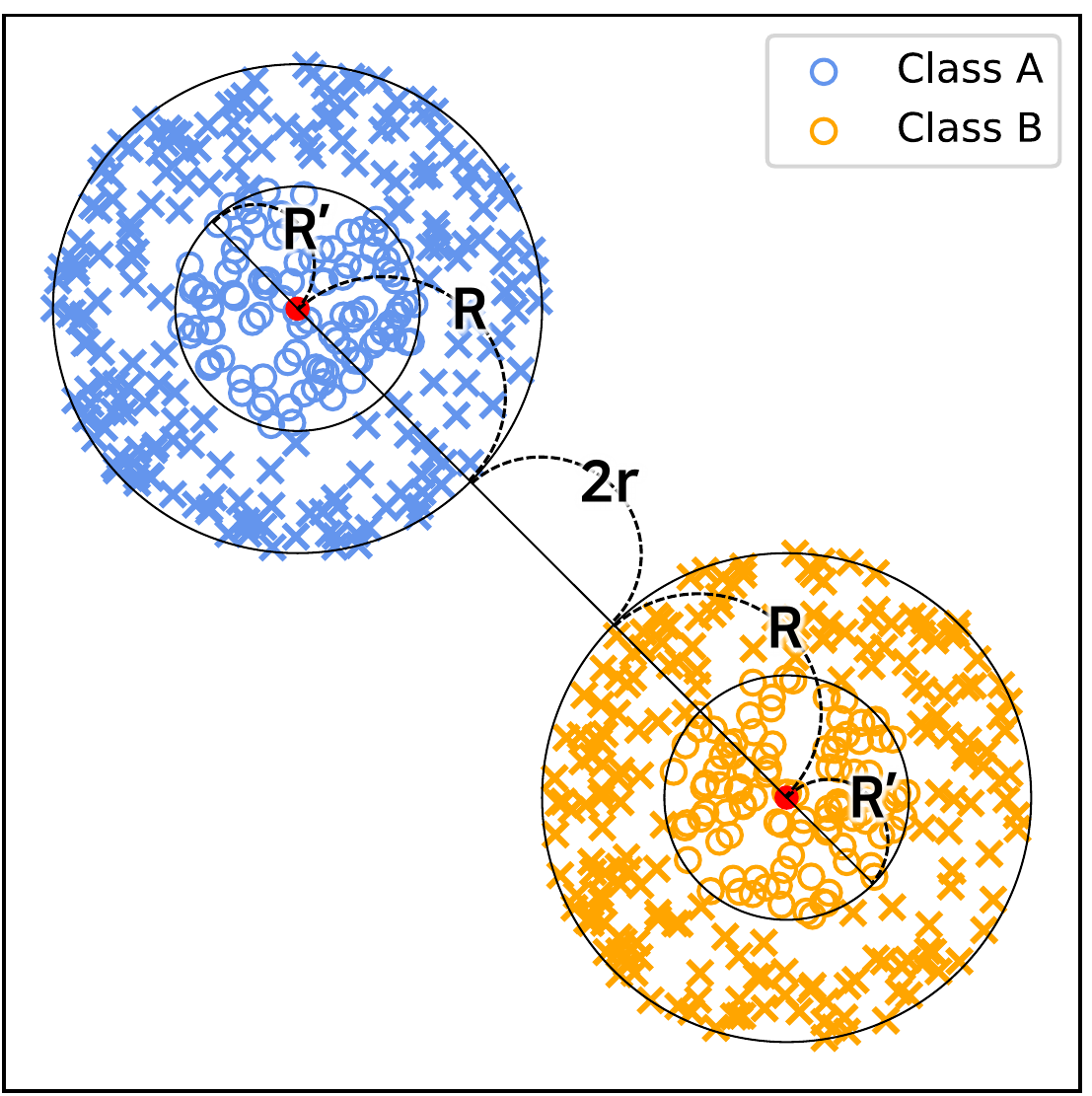}
    \label{fig:introduction:alignment}
  }
  \caption{
  Visual illustrations of data distribution manifolds with colors indicating sample labels.
  (a) Illustrates the robustness-accuracy tradeoff problem regarding standard clean accuracy and robust accuracy.
  (b) Shows misaligned distribution where test sample 'x' might differ in color from its nearest training sample 'o', despite large class distances. 
  (c) Depicts aligned distribution satisfying both separation and clustering, ensuring test sample 'x' matches the color of training sample 'o' if the minimum class distance is at least twice the radius of 'o'.
  }
  \label{fig:introduction}
\end{figure*}

Deep Neural Networks (DNNs), despite their high accuracy on clean samples, are notably susceptible to adversarial examples. These examples, involving test inputs with imperceptible perturbations, can lead to misclassifications, posing significant concerns in safety-critical domains such as autonomous driving and medical diagnosis~\cite{szegedy2013intriguing, papernot2016limitations, dong2018boosting, moosavi2016deepfool, carlini2017towards, kurakin2016adversarial}. Adversarial training has emerged as a crucial defensive technique over the past decade, enhancing the security of deep learning systems against such adversarial threats~\cite{goodfellow2014explaining,madry2017towards,tramer2017ensemble,kurakin2016adversarial}. Unlike methods that rely on auxiliary models~\cite{meng2017magnet,ma2018characterizing}, adversarial training directly enhances a classifier's robustness. It focuses on learning robust parameters to minimize adversarial losses, increasingly being generalized and certified across various samples. Recent advancements have aimed to guarantee or raise the lower bound of DNN robustness against $\epsilon$-bounded adversarial perturbations~\cite{li2023sok,zhang2023diffsmooth,yuan2023precise}. Initially focused on image classification, the concept of robust training is now expanding into other areas, including federated learning~\cite{zizzo2020fat,zhang2023delving,li2023improve,reisizadeh2020robust} and malware detection~\cite{lucas2023adversarial}, marking a significant evolution in the field.

The primary goal of adversarial training in DNNs is to enhance robustness against adversarial examples. However, this often comes at the cost of reduced accuracy on clean samples, presenting a significant robustness-accuracy tradeoff~\cite{tsipras2018robustness,zhang2019theoretically}. This tradeoff is particularly problematic in real-world applications, where the prevalence of normal samples means a robust but less accurate model may underperform compared to a standard model. For instance, in anomaly detection tasks like malware or fraud detection, this could lead to an unacceptable increase in false positives, compromising the model's practical utility.
While the robustness-accuracy tradeoff was once thought unavoidable~\cite{tsipras2018robustness}, recent research has offered new perspectives. Yang et al.~\cite{yang2020closer} suggested that certain dataset characteristics, like class separation, might mitigate this tradeoff. However, our work argues that this property alone is insufficient, as shown in Figure~\ref{fig:introduction:separation}. We propose two new data distribution properties: \textit{clustering} and \textit{alignment}. Clustering refers to the proximity of samples within a class, while alignment combines separation and clustering. We demonstrate that misclassification is virtually eliminated in datasets that exhibit these properties. See Figure~\ref{fig:introduction:alignment}.

Our analysis of actual datasets reveals a critical finding: separation does not imply alignment, leading to potential misclassification risks. This suggests that the tradeoff issue is inherent in the input space, contrary to the finding of~\cite{yang2020closer}. Therefore, we propose that aligning the feature space, particularly at the penultimate layer output of neural networks, is crucial. Our findings indicate that existing training algorithms do not effectively align the feature space, leading to misclassification. This insight underscores the necessity of a training algorithm that aligns the feature space to resolve the tradeoff in neural networks.

We propose \textit{Adversarial Feature Alignment (AFA)}, a novel adversarial training method targeting DNN feature extractors to improve the robustness-accuracy tradeoff.
AFA aims to identify and minimize risks from adversarial examples causing feature space misalignment. 
Utilizing contrastive learning~\cite{chen2020simple,khosla2020supervised}, known for its efficacy in feature space generalization, AFA operates under three key principles: 
(1) using adversarial examples that lead to the most significant feature misalignment, 
(2) ensuring these examples adhere to the class-label data manifold, and 
(3) providing precise guidance for their alignment. 
We have developed a new fully-supervised contrastive loss function, \textit{the AFA loss}, that meets these criteria and is optimized through a min-max approach. 
Figure~\ref{fig:overview} illustrates the overview of our approach.
AFA uniquely creates adversarial examples that amplify feature distance from their true class while reducing it from other classes, targeting the worst-case scenario for separation and clustering. 
Unlike previous self-supervised adversarial contrastive learning methods that struggle with class collision\footnote{Current methods face class collision problems~\cite{zheng2021weakly} because their loss functions include same-class samples as negatives in the anchor set, which negatively impacts feature alignment.} problems~\cite{kim2020adversarial,jiang2020robust,ho2020contrastive,fan2021does,yu2022adversarial}, AFA excels in aligning features robustly across classes, significantly enhancing the security and practicality of deep neural network training. 
We propose two distinct training strategies using AFA loss: initially pre-training the feature extractor with AFA followed by fine-tuning the linear classifier, and alternatively, training the entire network through joint optimization of adversarial and AFA losses. 
The joint optimization of AFA further improves the state-of-the-art accuracy and robustness when combined with the recent approach~\cite{wang2023better} that leverages a diffusion model~\cite{karras2022elucidating} for data augmentation.
Our experimental results confirm that AFA outperforms existing methods in both accuracy and robustness, and its efficacy is further amplified when integrated with recent diffusion model-based data augmentation techniques.

\smallskip\noindent\textbf{Contributions.}
This paper makes the following key contributions:

\begin{itemize}

\item We offer a new approach to address the robustness-accuracy tradeoff, focusing on aligning the separated data distribution through clustering within each class. Contrary to previous beliefs, our experiments reveal that this tradeoff is inherent in the input space of real-world datasets, primarily due to misaligned feature representations in neural networks.

\item We introduce 'Adversarial Feature Alignment (AFA)', a novel robust pre-training method that aligns feature representations to resolve the tradeoff in neural networks. AFA uniquely employs adversarial supervised contrastive learning for the neural network's feature extractor, marking the first instance of applying fully supervised contrastive learning to the adversarial min-max problem.

\item In our experiments, AFA has demonstrated improved robustness over existing adversarial training methods while maintaining accuracy on natural samples. Our method successfully learns more distinct feature spaces and smoother decision boundaries during pre-training.

\end{itemize}

\smallskip\noindent\textbf{Orgnaizations.}
\S\ref{sec:background} covers the preliminary concepts and the threat model.
\S\ref{sec:alignment} represents our key findings on alignment and the misalignment problem.
\S\ref{sec:afa} details the training strategy of AFA.
\S\ref{sec:evaluation} evaluates AFA's performance.
\S\ref{sec:discussion} discusses the implications of our work.
\S\ref{sec:related_work} review related work.
\S\ref{sec:conclusion} concludes this paper.
Supplementary materials, such as proofs and experiment details, can be found in the Appendices (\S\ref{sup:organization}$\sim$\S\ref{sup:experiments}).
\section{Background} \label{sec:background}

\subsection{Preliminary Concept} \label{sec:background:preliminary}

\smallskip\noindent\textbf{Deep neural network.}
Let a function $f: \mathcal{X} \rightarrow \mathbb{R}^N$ that maps input data $\mathcal{X}\subset \mathbb{R}^M$ into the prediction probabilities, where $M$ is the input dimensionality, and $N$ is the number of classes. Let $f(x)_i$ denote the probability of the $i$-th class for $i\in[N]$.
Partition $\mathcal{X}$ into the training set $\mathcal{X}_{\text{train}}$ and test set $\mathcal{X}_{\text{test}}$ where $\mathcal{X}_{\text{train}}\cup\mathcal{X}_{\text{test}}=\mathcal{X}$ and $\mathcal{X}_{\text{train}}\cap\mathcal{X}_{\text{test}}=\varnothing$.

We denote the neural network as a function $f_\text{dnn}$. We separate the function $f_\text{dnn}$ into the $L$-layer feature extractor $g$ and the linear classifier $h$ such that $f_\text{dnn}=h \circ g=h(g(x))$. $g_l$ is the $l$-th layer output of $g$ for $l \in \{1, ..., L\}$, and $g(x)=g_L(x)$ is identical to the output of the penultimate layer of $f_\text{dnn}$. $h$ estimates the importance of each class using the represented feature $g(x)$.
Finally, a classifier $F$ determines the predicted class of $x$ by $F(x)=\underset{i}{\text{argmax}}f(x)_i$.
Our scope in this paper is image classification task and convolution layer-based neural networks.

\smallskip\noindent\textbf{Definition of robustness and accuracy.}
Clean accuracy is the probability that the prediction of a classifier $F$ for input $x$ in the data distribution $\mu$ is identical to $y\in\mathcal{Y}$, the true class of a clean sample $x$ (i.e., $\underset{(\mathcal{X},\mathcal{Y})\sim\mu}{\text{Pr}}[F(x)=y$ for all $x\in\mathcal{X}]$).
Let $\mathbb{B}(x,\epsilon)$ denote a ball of radius $\epsilon>0$ around a sample $x\in\mathcal{X}$.
Robustness is the probability that the prediction of $F$ for all $x'\in\mathbb{B}(x,\epsilon)$ is identical to the prediction of $F$ for the original input $x$ (i.e., $\underset{(x,y)\sim\mu}{\text{Pr}}[F(x')=F(x)$ for all $x'\in\mathbb{B}(x,\epsilon)$).
Further, astuteness~\cite{wang2018analyzing,yang2020closer} is the probability that the prediction of $F$ for all $x'\in\mathbb{B}(x,\epsilon)$ is identical to the label $y$ (i.e., $\underset{(x,y)\sim\mu}{\text{Pr}}[F(x')=y$ for all $x'\in\mathbb{B}(x,\epsilon)$).
That is, astuteness can be used as robust accuracy for adversarial examples. For the rest of this paper, we refer to accuracy as the integration of clean and robust accuracies.

\subsection{Threat Model} \label{sec:background:threat}

\subsubsection{Adversarial attack}
The adversary in this paper performs an evasion attack that causes the misprediction of DNN. Given an original input $x$ and its true class $y$, the adversarial objective is generating an adversarial example $x'$ that satisfies:
\begin{center}
$\text{minimize}\ \|\delta\|_p\text{, such that}$\\
$F(x')\neq y\text{, where}\ x'=x+\delta$.
\end{center}
This is the untargeted attack to make DNN misclassify $x$ into any other class than $y$. The adversarial objective is changed into $F(x')=t$ for the targeted attack, where $t\neq y$ is the target class. Adversarial perturbation $\delta$ means the distortion applied to $x$. The perturbation is generated by propagating the gradient of the loss on the adversarial objective to the input layer. The optimization method for $\delta$ differs for the type of attack. $L_1$ and $L_2$ adversaries incorporates the regularization on the distortion in the loss function. $L_0$ and $L_\infty$ adversaries limits the size of perturbation within $\epsilon$. The representative adversarial evasion attack algorithms are fast gradient sign method (FGSM)~\cite{goodfellow2014explaining}, projected gradient descent (PGD)~\cite{madry2017towards}, Carlini and Wagner (CW) attack~\cite{carlini2017towards}, and AutoAttack (AA)~\cite{croce2020reliable}.

\subsubsection{Adversarial training}
The defender in this paper uses adversarial training. The primary objective of this defense method is training a DNN to construct its parameters that correctly classifies as many adversarial examples as possible (i.e., maximize the robust accuracy). To accomplish this, adversarial training finds worst-case perturbations and minimizes the risk of the perturbations on the model for each training batch. It is formulated as the following saddle point problem~\cite{madry2017towards}:
\begin{equation}
\label{eq:saddle_point}
    \underset{\theta}{\min}\ \mathbb{E}_{(x, y)\sim \mathcal{D}} \underset{||\delta||\leq\epsilon}{\max} \mathcal{L}_{CE}(x+\delta, y; \theta).
\end{equation}

The equation above can be considered as a kind of empirical risk minimization (ERM). Eq.~\ref{eq:saddle_point} jointly optimizes the inner maximization and outer minimization problems. The inner optimization problem seeks a perturbation $\delta$, within the radius $\epsilon>0$, that maximizes the cross entropy loss for given input $x\in \mathcal{X}$ and its class label $y\in \mathcal{Y}$ on a data distribution $\mathcal{D}$. In \cite{madry2017towards}, the loss is maximized by the projected gradient descent. The outer optimization problem updates the network parameter $\theta$ such that the adversarial loss of $x+\delta$ is minimized.

We present the additional requirement for adversarial training that its accuracy on clean samples should be preserved. This is very important to guarantee the availability of the model. In the deep learning environment, most samples that the model addresses are clean samples, and adversarial examples are relatively rare. In this respect, the total number of inputs that the deep learning system can handle decreases as the clean accuracy degrades, even if the robust accuracy is high enough.
The robustness-accuracy tradeoff should be solved for the practicality of adversarial training in real-world applications.
\section{Robustness and Accuracy Need Alignment} \label{sec:alignment}

\begin{table*}
\begin{footnotesize}
\begin{center}
\caption{Separation and clustering factors and the accuracy of the 1-nearest neighbor $f_\text{1-nn}$ for the input space of various datasets. The separation factor $2r$ is the minimum/maximum distance between two different classes. The clustering factor $2R$ is the minimum/average/maximum distance of samples within the same class.
The column "Train-Train" measures the distance between samples within the same training dataset. The column "Train-Test" measures the distance between training and clean test samples.  
Pixel values are normalized to the range [0, 1]. We employed $l_\infty$ based on~\cite{yang2020closer}, positing that datasets are distinct since the $l_\infty$ separation factor in the input space exceeds the $l_\infty$ adversarial perturbation size. Results for other distance metrics are reported in Table~\ref{tab:sup:alignment:input_space} in Appendix~\ref{sup:alignment:input_space}.
}
\label{tab:alignment:input_space}
\renewcommand{\arraystretch}{1.4}
\begin{tabular}{l||M{0.7cm}M{0.7cm}M{0.7cm}|M{0.7cm}M{0.7cm}M{0.7cm}|M{0.7cm}M{0.7cm}|M{0.7cm}M{0.7cm}|M{2.5cm}}
\toprule
 & \multicolumn{6}{c|}{\textbf{Separation Factor}} & \multicolumn{4}{c|}{\textbf{Clustering Factor}} & \multirow{3}{*}{\textbf{Test Accuracy of $f_\text{1-nn}$}} \\
\textbf{Dataset} & \multicolumn{3}{c}{\textbf{Train-Train}} & \multicolumn{3}{c|}{\textbf{Train-Test}} & \multicolumn{2}{c}{\textbf{Train-Train}} & \multicolumn{2}{c|}{\textbf{Train-Test}} & \\
 & \textbf{Min} & \textbf{Avg} & \textbf{Max} & \textbf{Min} & \textbf{Avg} & \textbf{Max} & \textbf{Min} & \textbf{Max} & \textbf{Min} & \textbf{Max} & \textbf{(\%)}\\ \midrule[0.2ex]
MNIST           & 0.737 & 0.927 & 0.988 & 0.812 & 0.958 & 0.988 &   1.0 &   1.0 &   1.0 &   1.0 & 21.08 \\
CIFAR10         & 0.211 & 0.309 & 0.412 & 0.220 & 0.331 & 0.443 &   1.0 &   1.0 &   1.0 &   1.0 & 81.77 \\
CIFAR100        & 0.067 & 0.371 & 0.561 & 0.114 & 0.414 & 0.604 &   1.0 &   1.0 &   1.0 &   1.0 & 95.47 \\
STL10           & 0.369 & 0.493 & 0.624 & 0.345 & 0.466 & 0.627 &   1.0 &   1.0 &   1.0 &   1.0 & 88.66 \\
Restricted ImageNet     & 0.235 & 0.332 & 0.426 & 0.271 & 0.410 & 0.533 &   1.0 &   1.0 &   1.0 &   1.0 & 78.85 \\
\bottomrule
\end{tabular}
\end{center}
\end{footnotesize}
\end{table*}

\subsection{Properties of Data Manifold} \label{sec:alignment:property}

\smallskip\noindent\textbf{Separation.}
Yang et al.~\cite{yang2020closer} defined the separation property as a requirement of the data distribution for the astute classifier. Let $\mathcal{X}$ contain $N$ disjoint classes $\mathcal{X}^{(1)},...,\mathcal{X}^{(N)}$, where all samples in $\mathcal{X}^{(i)}$ have label $i$ for $i\in[N]$.

\begin{definition} [$r$-separation~\cite{yang2020closer}]
\label{def:separation}
    \textit{Let ($\mathcal{X}$, $\mathsf{dist}$) be a metric space. A data distribution over $\bigcup_{i\in [N]}\mathcal{X}^{(i)}$ is $r$-separated in the input space if $\mathsf{dist}(\mathcal{X}^{(i)},\mathcal{X}^{(j)})\geq 2r$ for all $i\neq j$, where $\{i, j\}\subset [N]$ and $\mathsf{dist}(\mathcal{X}^{(i)},\mathcal{X}^{(j)})=\textrm{min}_{x\in \mathcal{X}^{(i)},x'\in \mathcal{X}^{(j)}}\mathsf{dist}(x,x')$.}
\end{definition}

Definition~\ref{def:separation} indicates that the minimal distance between two different classes is larger than $2r$. As shown in~\cite{yang2020closer}, this property held for actual image datasets (e.g., MNIST, CIFAR10), and even $r$ was several times larger than the standard perturbation budget $\epsilon$.

\smallskip\noindent\textbf{Clustering.} We raise a possible case that a sample in the separated data distribution is quite far from samples in its true class even though it is far from other classes. We define a new property for data distribution, \textit{clustering}, to prevent this phenomenon.

\begin{definition} [$R$-clustering]
\label{def:clustering}
    \textit{We say that a data distribution over $\bigcup_{i\in[N]}\mathcal{X}^{(i)}$ is $R$-clustered if $\mathsf{dist}_\text{max}(x,\mathcal{X}^{(i)})\leq 2R$ for all $x\in\mathcal{X}^{(i)}$ and $i\in[N]$, where $\mathsf{dist}_\text{max}(x,\mathcal{X}^{(i)})=\text{max}_{x'\in\mathcal{X}^{(i)}\setminus x}\mathsf{dist}(x,x')$.}
\end{definition}

$r$-separation property only observes whether a sample exists in the different class of $x$ within radius $r$ around $x$, so it only guarantees the data manifold around $x$. On the other hand, $R$-clustering observes whether all samples in the same class reside in radius $R$, which guarantees the entire data manifold of a subset for the same class in the dataset.

\smallskip\noindent\textbf{Alignment.}
As shown in Figure~\ref{fig:introduction:alignment}, the data distribution can simultaneously satisfy separation and clustering. We define \textit{alignment} by integrating these two properties.

\begin{definition} [$R$-alignment]
\label{def:alignment}
    \textit{We say that a data distribution over $\bigcup_{i\in[N]}\mathcal{X}^{(i)}$ is $R$-aligned if the distribution satisfies $r$-separated and $R$-clustered for $r>R$.}
\end{definition}

If a distribution does not satisfy $R$-alignment, we say that the distribution has a \textit{misalignment problem}, where a sample is closer to another sample from a different class than the true class.

\smallskip\noindent\textbf{Nearest neighbor classifier.}
We start by observing the robustness-accuracy tradeoff of the 1-nearest neighbor classifier in Section~\ref{sec:alignment:input}.
Let $\mathsf{dist}(x,\mathcal{X}_{\text{train}}^{(i)})=\text{min}_{x_\text{ref}\in\mathcal{X}_{\text{train}}^{(i)}\setminus x}\mathsf{dist}(x,x_\text{ref})$ where $x\in\mathcal{X}$ and $\mathsf{dist}$ is a distance metric.
We define a 1-nearest neighbor (1-NN) binary classifier $f_\text{1-nn}$ with radius $R$ as follows:
\begin{equation}
\label{eq:nearest_neighbor_f_binary}
    f_\text{1-nn}(x)=\frac{1}{2R}\cdot\left(\mathsf{dist}(x,\mathcal{X}_{\text{train}}^{(0)}), \mathsf{dist}(x,\mathcal{X}_{\text{train}}^{(1)})\right).
\end{equation}

In this case, we let $F_\text{1-nn}=\text{argmin}(f_\text{1-nn}^{(i)})$ such that the predicted class of the test input $x$ is identical to the class of the nearest training sample of $x$. We note that there is no identical test sample $x$ as any training sample (i.e., $\mathsf{dist}(x,x_\text{ref})>0$).

\subsection{Separation Is Not Enough: Alignment Helps} \label{sec:alignment:input}

The separation property ensures robustness within a radius $r$ around a reference point $x_\text{ref}$. However, separation alone doesn't guarantee complete robustness and accuracy. This limitation arises because separation doesn't necessarily mean a test sample will be close to $x_\text{ref}$. For instance, as depicted in Figure~\ref{fig:introduction:separation}, even with separated data distributions, the nearest training sample to a test input might belong to a different class, leading to potential misclassifications. This issue can affect clean samples as well, thereby reducing clean accuracy. In Lemma~\ref{lem:accuracy_alignment}, we show that for complete accuracy, a separated dataset also needs to be clustered, or in other words, aligned.

\begin{lemma}
\label{lem:accuracy_alignment}
    $f_\text{1-nn}$ has accuracy of 1 on the $R$-aligned data distribution.
\end{lemma}

Accuracy is affected more by the density of samples of the same class than by their distance from other classes. As shown in Theorem~\ref{theo:train_alignment}, data distributions with training samples densely clustered around the center do not require more robust separation.

\begin{theorem}
\label{theo:train_alignment}
    Let a data distribution $\mathcal{X}$ is $R$-clustered and $\mathcal{X}_\text{train}\subset \mathcal{X}$ is $R'$-clustered around the center of $R$-ball with $R'\leq R$. Then $f_\text{1-nn}$ has accuracy of 1 on $\mathcal{X}$ if the distribution is $r$-separated with $r>R'$.
\end{theorem}

Proof of Lemma~\ref{lem:accuracy_alignment} and Theorem~\ref{theo:train_alignment} are in Appendix~\ref{sup:proof}.
When the training samples are clustered within a certain radius, as shown in Figure~\ref{fig:introduction:alignment}, if the test samples of two classes are spaced apart by no more than the diameter of the training samples, then the true classes of one test sample and its nearest neighbor are always the same.
Theorem~\ref{theo:train_alignment} states that the maximum radius of the training samples determines the minimum distance between different classes for complete accuracy. Therefore, the closer together training samples of the same class are, the better for accuracy. We also note that datasets with a larger minimum distance between classes are preferred because the model is more confident with the larger margin of the decision boundary.
Theorem~\ref{theo:train_alignment} is introduced to elucidate the tradeoffs in practical DL situations where the test data distribution can be more sparse than its training dataset. Given the challenge of optimally balancing this tradeoff, our insight is that by densifying the training samples, as in Theorem~\ref{theo:train_alignment}, the model gains robustness against test-phase outliers.

We identified whether the input space of datasets irrelevant to the model is aligned. In Table~\ref{tab:alignment:input_space}, the minimum distance of two different classes (separation factor) is large enough to exceed the perturbation budget $\epsilon$. However, we observed the misalignment problem: the maximum distance within a class was much larger than the separation factor. This result indicates that the samples in each class are not closely distributed, and there may be test samples closer to training samples in other classes than training samples in the ground-truth class. The misalignment problem of the input space makes the error rate of $f_\text{1-nn}$ based on $l_\infty$ distances between pixels considerably high. This problem also arises for other metrics, as shown in Table~\ref{tab:sup:alignment:input_space}. Thus, contrary to the claim of Yang et al.~\cite{yang2020closer}, the robustness-accuracy tradeoff is still intrinsic to the input space of neural networks.
\subsection{Misalignment Problem in Feature Space} \label{sec:finding:misalignment}

\begin{table*}
\begin{footnotesize}
\begin{center}
\caption{Separation and clustering factors for penultimate layer output of a neural network trained by different methods. The target dataset is CIFAR10, and the model architecture is ResNet-18 which was not pre-trained before applying each method. The column "Train-Test" measures the distance between training and clean test samples. The column "Train-Adv" measures the distance between training samples and PGD adversarial examples generated from clean test samples with $\epsilon=8/255$, step size $\alpha=2/255$, and attack iteration $N=20$. 
Feature values are normalized to the range [0, 1]. For the unconstrained feature space, we opted for the $l_1$ norm as the distance metric.
}
\label{tab:alignment:feature_space:cifar10}
\renewcommand{\arraystretch}{1.4}
\begin{tabular}{l||M{0.7cm}M{0.7cm}|M{0.7cm}M{0.7cm}|M{0.7cm}M{0.7cm}|M{0.7cm}M{0.7cm}|M{0.7cm}M{0.7cm}M{0.7cm}M{0.7cm}}
\toprule
 & \multicolumn{4}{c|}{\textbf{Separation Factor}} & \multicolumn{4}{c|}{\textbf{Clustering Factor}} & \multicolumn{4}{c}{\textbf{Test Accuracy (\%)}} \\
& \multicolumn{2}{c}{\textbf{Train-Test}} & \multicolumn{2}{c|}{\textbf{Train-Adv}} & \multicolumn{2}{c}{\textbf{Train-Test}} & \multicolumn{2}{c|}{\textbf{Train-Adv}} & \multicolumn{2}{c}{\textbf{Clean Samples}} & \multicolumn{2}{c}{\textbf{Adv. Examples}}\\
\textbf{Method}  & \textbf{Min} & \textbf{Max} & \textbf{Min} & \textbf{Max} & \textbf{Min} & \textbf{Max} & \textbf{Min} & \textbf{Max} & \textbf{$f_\text{1-nn}$} & \textbf{$f_\text{dnn}$} & \textbf{$f_\text{1-nn}$} & \textbf{$f_\text{dnn}$}\\ \midrule[0.2ex]
Cross-entropy
& 1.358 & 9.030 & 0.938 & 1.836 & 20.161 & 23.496 & 23.729 & 25.067 &  92.65 &  92.87 &      0 &      0 \\
PGD AT
& 0.970 & 7.958 & 1.480 & 5.340 & 20.729 & 22.659 & 22.860 & 24.632 &  81.80 &  83.77 &  47.82 &  49.18 \\
TRADES $\beta=1$
& 1.789 & 7.940 & 1.608 & 6.333 & 19.959 & 21.980 & 21.706 & 23.352 &  86.93 &  86.49 &  43.55 &  46.29 \\
TRADES $\beta=6$
& 1.516 & 6.855 & 1.460 & 5.496 & 21.820 & 22.743 & 22.343 & 23.355 &  81.15 &  80.84 &  49.26 &  51.23 \\
\bottomrule
\end{tabular}
\end{center}
\end{footnotesize}
\end{table*}

Since it is difficult to solve the tradeoff in the input space, we delve into the feature space of a deep neural network, which can generalize the data distribution through training.
Extending the perspective of~\cite{yang2020closer} and Section~\ref{sec:alignment:input}, we say that the feature space should be sufficiently separated and aligned for robustness and accuracy.
In this respect, we look at the relationship between the output of the linear classifier $h$ and the manifold of the extracted feature $g(x)$, which is the input space of $h$.
Although we cannot derive the exact radius that is tolerable by the linear classifier, we can estimate whether samples of different classes are located in the separated manifold.
In Table~\ref{tab:alignment:feature_space:cifar10}, we report the accuracies of the neural network and 1-nearest neighbor on the feature space of the network.
We measured the distance of feature vectors from input to training samples for $f_\text{1-nn}$.
As is well known, there was the robustness-accuracy tradeoff for $f_\text{dnn}$. Surprisingly, we identified the same tradeoff for $f_\text{1-nn}$. The similarity between the accuracies of the two functions demonstrates that the feature misalignment problem leads to the misclassification of the network.
From Figure~\ref{fig:alignment}, we identify that the accuracy of $f_\text{1-nn}$ on the penultimate layer is the most similar to that of the neural network. Our finding implies that Lemma~\ref{lem:accuracy_alignment} and Theorem~\ref{theo:train_alignment} hold in part for the linear classifier on the feature space. However, Table~\ref{tab:alignment:feature_space:cifar10} shows
that training methods result in the separation factor much larger than the clustering factor. In other words, a large part of the distribution of the two classes overlaps. As shown in Table~\ref{tab:sup:alignment:feature_space:cifar10} to Table~\ref{tab:sup:alignment:feature_space:res_imagenet} in Appendix~\ref{sup:alignment:feature_space}, this result was the same for other model architectures and datasets.
In addition, we observed misalignments in the feature space for other distance metrics through Table~\ref{tab:sup:alignment:feature_space:cifar10:metrics}; separation factors were lower than alignment factors. The $f_\text{1-nn}$ accuracy was consistent in feature space against PGD adversarial examples across metrics, implying metric-independence of the feature misalignment.

Our finding is similar to the previous research that measured the tradeoff between an error rate and a distance to the nearest neighbor in the concentric sphere~\cite{gilmer2018adversarial}. Our difference is that we observed the tradeoff between the error rate and distances among samples in the actual feature space. Table~\ref{tab:alignment:feature_space:cifar10} summarizes
that a neural network cannot generalize the feature representation with natural and adversarial training. Further, it requires a robust feature alignment method that clusters and separates classes in feature space to solve the tradeoff. 

\subsection{Feature Misalignment of Different Layers} \label{sec:finding:layer}

\begin{figure*}[t]
  \centering
   \subfigure[CIFAR10-ResNet-18]{
     \includegraphics[width = 0.3\textwidth]{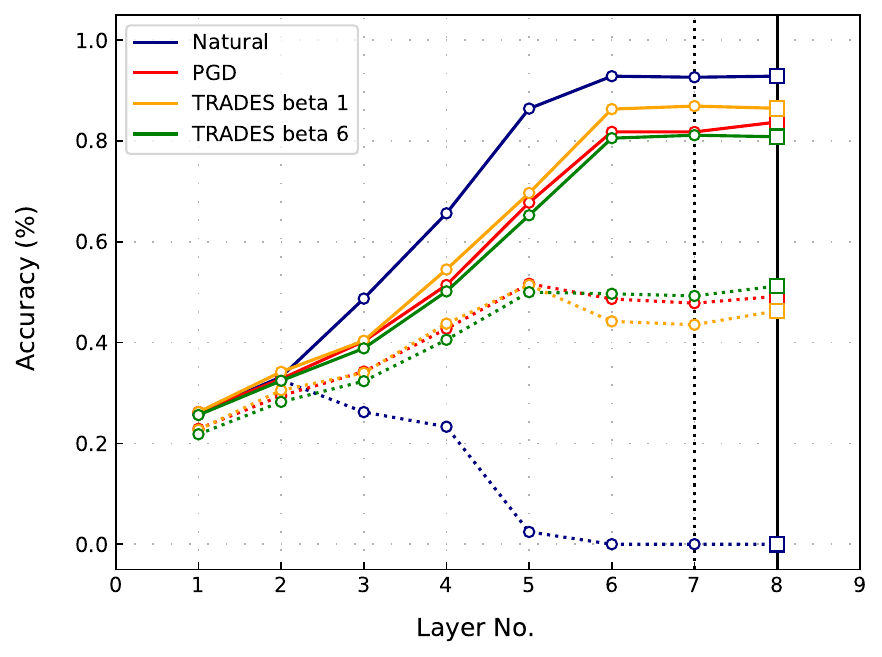}
     \label{fig:alignment:cifar10-res18}
   }
   \subfigure[CIFAR10-WideResNet-36-10]{
     \includegraphics[width = 0.3\textwidth]{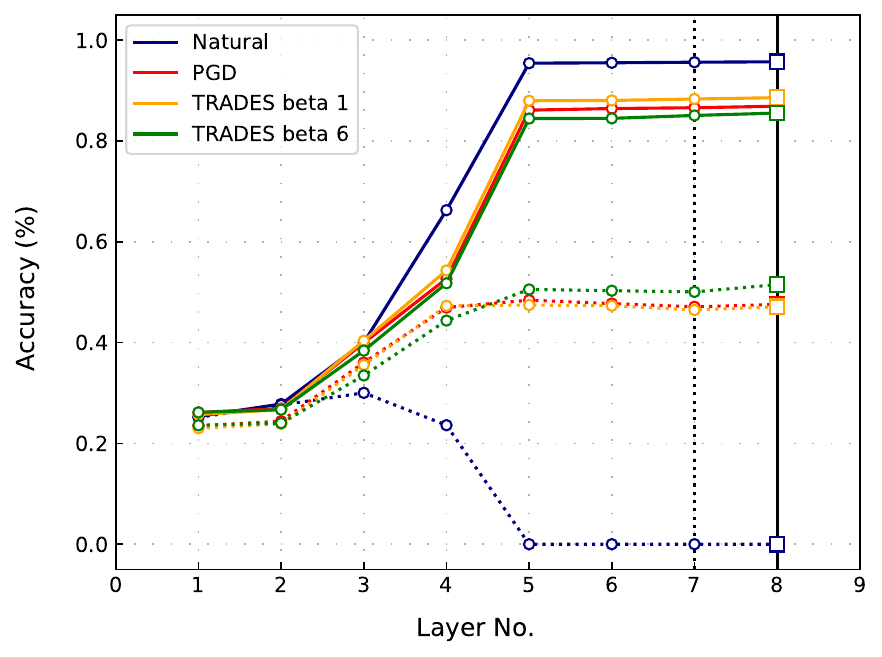}
     \label{fig:alignment:cifar10-wideres}
   }
   \subfigure[CIFAR100-ResNet-18]{
     \includegraphics[width = 0.3\textwidth]{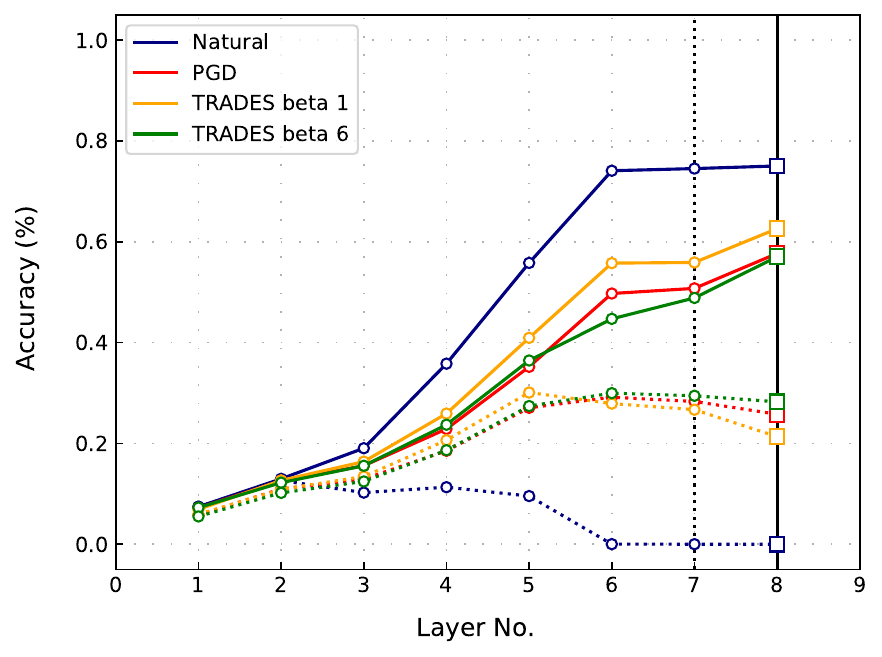}
     \label{fig:alignment:cifar100-res18}
   }
  \caption{Accuracy of clean samples (solid lines) and PGD adversarial examples (dashed lines) on each neural network layer with different training methods. Vertical dashed lines indicate the penultimate layer, and vertical solid lines indicate the logit layer. For layers before the logit layer (i.e., the last layer), the accuracy indicates the accuracy of the $f_\text{1-nn}$. The accuracy of the logit layer is identical to the classification accuracy of the network. PGD examples were generated from clean test samples with $\epsilon=8/255$, step size $\alpha=2/255$, and attack iteration $N=20$. We measured the $l_1$ distance between the layer output of a test sample and those of train samples.}
  \label{fig:alignment}
\end{figure*}

We observed the relationship between the alignment in the feature representation of hidden layers and the classification accuracy. Figure~\ref{fig:alignment} describes the accuracy of each neural network layer for test inputs.
The accuracy of layers before the penultimate layer denotes the accuracy of the 1-nearest neighbor $f_\text{1-nn}$ that uses the output of each layer.
We identified consistent results for all datasets, architectures, and training methods. The accuracy of the first layer in Figure~\ref{fig:alignment} cannot discriminate between clean samples and adversarial examples. As the layer went deeper, the accuracy converged toward that of the logit layer: the accuracy of the penultimate layer was the most similar to the logit layer. These results imply that the penultimate layer output is the most influential in the linear classifier.

\begin{table}[t]
\begin{footnotesize}
\begin{center}
\caption{Accordance rate of training methods for ResNet-18 on CIFAR10 and CIFAR100. The accordance rate is the ratio of samples the majority of whose k-nearest neighbors in feature space have the same class as the predicted class from the neural network for the samples. We used $\epsilon=8/255$, step size $\alpha=2/255$, and attack iteration $N=20$ for PGD attack. We used the $l_1$ distance between the penultimate layer output of a test sample and those of training samples.}
\label{tab:sup:misalignment}
\renewcommand{\arraystretch}{1.4}
\begin{tabular}{l||M{0.9cm}M{0.9cm}|M{0.9cm}M{0.9cm}}
\toprule
\multirow{2}{2.4cm}{\textbf{Training Method}} & \multicolumn{2}{c|}{\textbf{Clean (\%)}} & \multicolumn{2}{c}{\textbf{PGD(\%)}}\\
& \textbf{k=1} & \textbf{k=100} & \textbf{k=1} & \textbf{k=100} \\ \midrule[0.2ex]
\rowcolor{gray!20}\multicolumn{5}{c}{CIFAR10 Dataset} \\
Cross-entropy                   & 96.54 & 98.39 & 99.99 & 99.99 \\
PGD AT~\cite{madry2017towards}  & 76.56 & 87.74 & 67.95 & 83.04 \\
AdvCL~\cite{fan2021does}        & 84.46 & 89.46 & 76.16 & 82.98 \\
AFA (ours)                      & 95.34 & 96.07 & 84.43 & 87.95 \\
\rowcolor{gray!20}\multicolumn{5}{c}{CIFAR100 Dataset} \\
Cross-entropy                   & 74.83 & 81.02 & 96.27 & 98.03 \\
PGD AT~\cite{madry2017towards}  & 48.18 & 63.73 & 40.65 & 60.19 \\
AdvCL~\cite{fan2021does}        & 53.93 & 69.81 & 46.49 & 65.82 \\
AFA (ours)                      & 70.72 & 76.12 & 53.00 & 63.62 \\
\bottomrule
\end{tabular}
\end{center}
\end{footnotesize}
\end{table}

\subsection{Correlation between Misaligned Classes and Misclassified Classes} \label{sec:finding:accordance}

We further observe whether a class whose manifold the feature of a sample is located affects the prediction of the linear classifier. We deploy a k-nearest neighbor (k-NN) classifier that uses the outputs of the feature extractor as an input. We identify whether the predicted class of the linear classifier corresponds with the majority of k-nearest neighbors. We set k to 1 and 100.

The accordance rate in Table~\ref{tab:sup:misalignment} describes the ratio of samples that the neural network and k-NN classifier yield the same class. With natural cross-entropy, predictions of the network and the k-NN classifier agreed for most samples regardless of the misclassification. One possible reason is that natural training uses only clean samples to learn feature distribution, so the feature representation of the network becomes monotonous. This result indicates that the manifold where the feature of a sample locates is significantly utilized for the decision of the linear classifier. Adversarial training methods showed lower accordance rates than natural training. Nevertheless, their accordance rates are still high, implying that linear classifiers from adversarial training depend on the manifold in the feature space. We also see that our method yields the highest accordance rates among adversarial training methods. From our method, samples are better aligned along their classes in the feature space, and the uncertainty of the linear classifier is mitigated.
\section{Adversarial Feature Alignment} \label{sec:afa}

\begin{figure*}[t]
  \centering
  \subfigure[Inner optimization step]{
    \includegraphics[width = 0.3\textwidth]{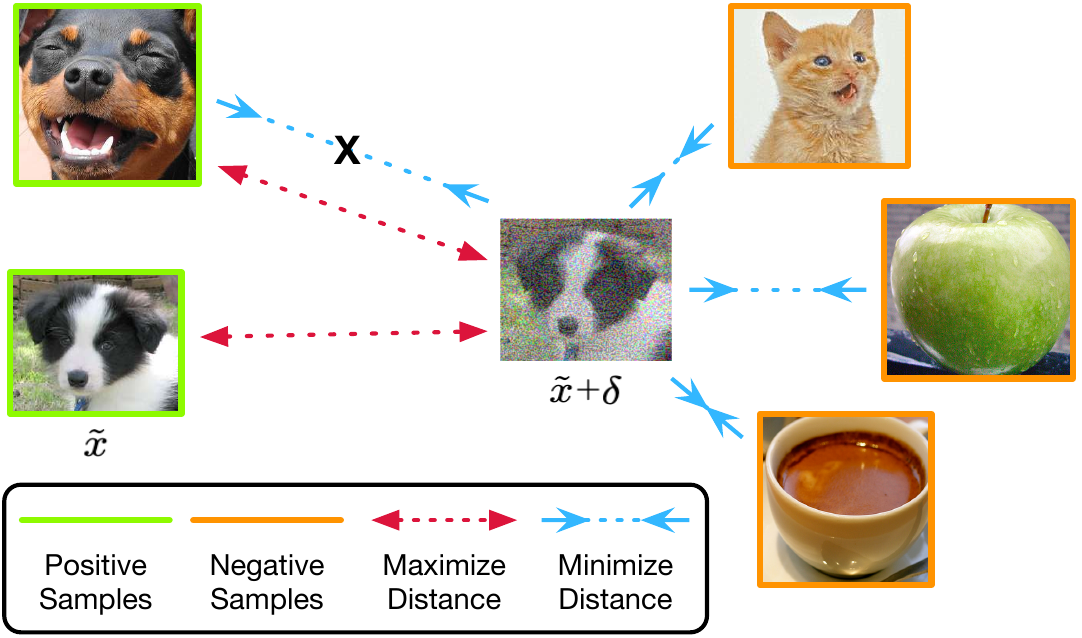}
    \label{fig:overview:inner}
  }
  \centering
  \subfigure[Outer optimization step]{
    \includegraphics[width = 0.3\textwidth]{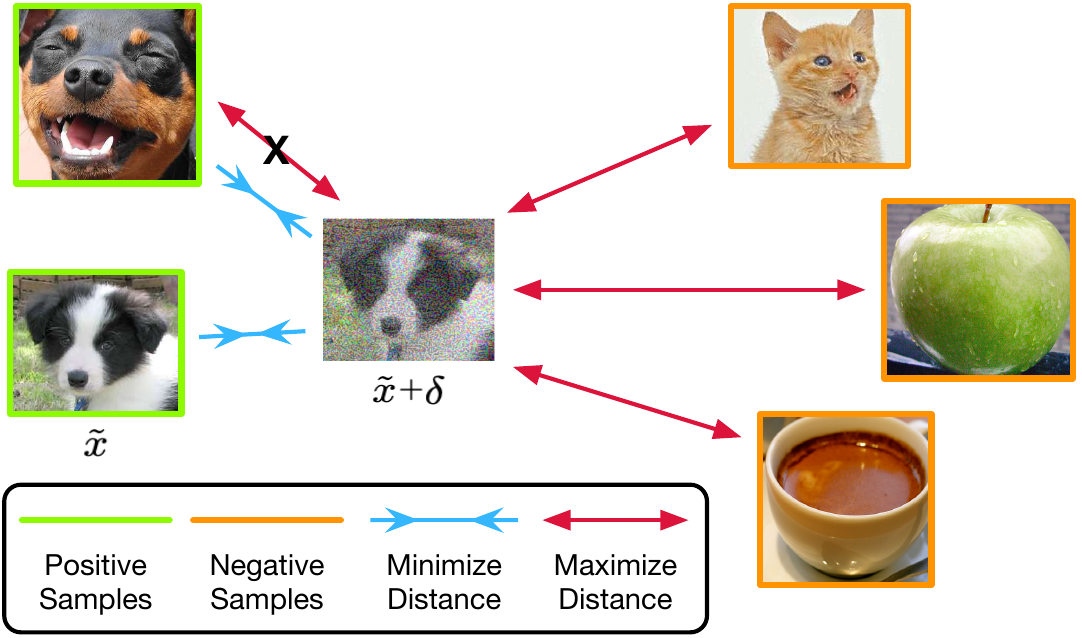}
    \label{fig:overview:outer}
  }
  \centering
  \subfigure[After training]{
    \includegraphics[width = 0.3\textwidth]{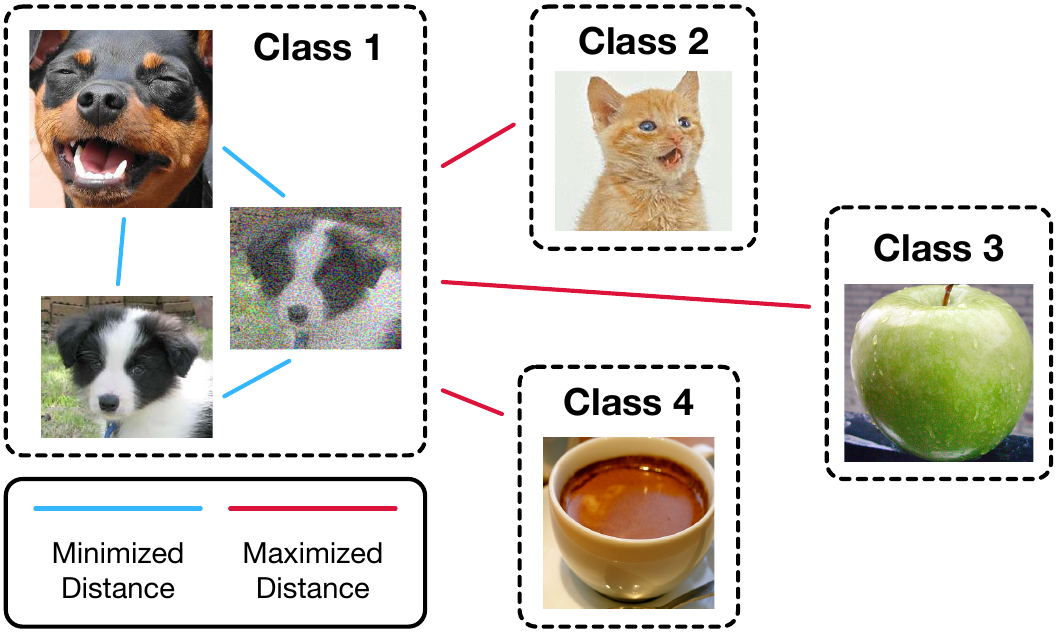}
    \label{fig:overview:after}
  }
  \caption{Overview of Adversarial Feature Alignment (AFA): AFA incorporates inner and outer optimization steps in each training epoch. (a) In the inner step, an adversarial example $\tilde{x}+\delta$ is generated to maximize the AFA adversarial loss, optimizing feature vector distances from both positive and negative examples. (b) The feature extractor $g$ is then updated to minimize the feature vector distance from positive examples and maximize it from negatives. (c) Post-AFA training, class samples are efficiently clustered into their respective classes, optimizing both intra-cluster closeness and inter-cluster separation.
  }
  \label{fig:overview}
\end{figure*}

In this section, we propose Adversarial Feature Alignment (AFA), a new robust training method for the feature extractor $g$ of $f_\text{dnn}$. We define a new adversarial contrastive loss in a fully-supervised manner and solve the min-max optimization problem that targets the loss, as illustrated in Figure~\ref{fig:overview}. AFA effectively solves the robustness-accuracy tradeoff because it finds the worst-case sample that degrades alignment and optimizes it.
In this respect, we revisit the existing contrastive loss function in \S\ref{sec:afa:revisit} and discuss principles of a new loss function for the effectiveness of AFA in \S\ref{sec:afa:requirement}. We design AFA loss function in \S\ref{sec:afa:loss} and new optimization strategy of AFA in \S\ref{sec:afa:strategy}.

\subsection{Revisiting Contrastive Loss Function} \label{sec:afa:revisit}

Contrastive learning is a training method to pretrain the feature extractor of a DNN. The objective is minimizing the feature distance between an anchor sample and its positive samples and maximizing the feature distance between the anchor and its negatives.
Let $X=\{x_1, x_2, ..., x_n\}$ be an input batch, and $Y=\{y_1, y_2, ..., y_n\}$ be its corresponding class label batch. The supervised contrastive loss is measured over a multiview batch. In this paper, we consider the multiview batch $\tilde{X}=\{\tilde{x}_1, \tilde{x}_2, ..., \tilde{x}_{kn}\}$, which is the $k$-fold augmentation of $X$, where $\{\tilde{x}_{k(i-1)+1}, ..., \tilde{x}_{ki}\}$ are randomly transformed images of $x_i$. $Y$ is also augmented to $\tilde{Y}=\{\tilde{y}_1, \tilde{y}_2, ..., \tilde{y}_{kn}\}$, where $\tilde{y}_{k(i-1)+1}=...=\tilde{y}_{ki}=y_i$.

\smallskip\noindent\textbf{Self-supervised contrastive loss function.}
Self-supervised contrastive learning~\cite{chen2020simple} does not use the class label. Instead, it labels the augmentations of the same original sample as the anchor as positive and the remaining samples in the batch as negative. Samples derived from original samples different than the anchor are labeled negative, even though their class is the same as the anchor.
The self-supervised contrastive loss $\mathcal{L}_{self}$ is formulated as follows:
\begin{equation}
\label{eq:selfcon}
    \mathcal{L}_{self} = \sum_{i=1}^{kn} \frac{-1}{|P_{self}(\tilde{x}_i)|} \sum_{\tilde{x}_p\in P_{self}(\tilde{x}_i)} \log\frac{e^{z_{\tilde{x}_i}\cdot z_{\tilde{x}_p}/\tau}}{\underset{a\in A(\tilde{x}_i)}\sum_{} e^{z_{\tilde{x}_i}\cdot z_a/\tau}}.
\end{equation}

In Eq.~\ref{eq:selfcon}, $z_x = \phi(g(x)) \in \mathbb{R}^{D_P}$ is a normalized feature embedding of $g(x)$ by the projection layer $\phi$. $\phi$ is a multi-layer perceptron and is used only for the pre-training phase.
$A(\tilde{x}_i)=\tilde{X}\ \backslash\ \{\tilde{x}_i\}$ is the contrastive view of the anchor sample $\tilde{x}_i$ and incorporates positive and negative samples of $\tilde{x}_i$.
$P_{self}(\tilde{x}_i)$ is a set of positive samples of $\tilde{x}_i$. It consists of samples derived from the original sample same as $\tilde{x}_i$.
Negative samples of $\tilde{x}_i$ in $\mathcal{L}_{self}$ are $A(\tilde{x}_i) \backslash P_{self}(\tilde{x}_i)$.
The inner dot product calculates the distance of feature embeddings between two samples. $\tau$ is a temperature to regularize inner dot products.
In summary, for each training sample, the loss function in Eq.~\ref{eq:selfcon} calculates the feature product with the positives that should be maximized in the numerator and the feature product with the negatives that should be minimized in the denominator.

\smallskip\noindent\textbf{Supervised contrastive loss function.}
It is known that the fully-supervised contrastive learning~\cite{khosla2020supervised} performs better than the self-supervised one~\cite{chen2020simple} in general.
This approach has a different criterion on positive and negative samples from the self-supervised one: all samples in the same class as the anchor are positive and samples in the other classes are negative.
The supervised contrastive loss $\mathcal{L}_{sup}$ is formulated as follows:
\begin{equation}
\label{eq:supcon}
    \mathcal{L}_{sup} = \sum_{i=1}^{kn} \frac{-1}{|P_{sup}(\tilde{x}_i)|} \sum_{\tilde{x}_p\in P_{sup}(\tilde{x}_i)} \log\frac{e^{z_{\tilde{x}_i}\cdot z_{\tilde{x}_p}/\tau}}{\underset{a\in A(\tilde{x}_i)}\sum_{} e^{z_{\tilde{x}_i}\cdot z_a/\tau}}.
\end{equation}

Eq.~\ref{eq:supcon} and Eq.~\ref{eq:selfcon} are almost the same but the difference is the term for a positive set. The positive set in Eq.~\ref{eq:supcon} is $P(\tilde{x}_i):=\{\tilde{x}_p\in A(\tilde{x}_i) : \tilde{y}_p=\tilde{y}_i\}$.
$L_{sup}$ of Eq.~\ref{eq:supcon} is the same as $L^{sup}_{out}$ of~\cite{khosla2020supervised}. $L^{SCL}_{out}$, whose summation over positives of an anchor is outside the log, had better performance than $L^{sup}_{in}$ in that work.

\subsection{Principles for Adversarial Feature Alignment} \label{sec:afa:requirement}

\smallskip\noindent\textbf{Adversarial training over contrastive loss is necessary.}
Vanilla contrastive learning is the simple alignment of clean samples, so the contrastive loss does not address potentially misaligned samples, even if the clean samples can be aligned in feature space. This means that the model is overfitted to the manifold of clean samples and is vulnerable to adversarial examples that force misalignment.
Therefore, we need to maximize the risk of feature misalignment during the learning process.

\smallskip\noindent\textbf{Supervised contrastive loss better than self-supervised one.}
The inner step of the self-supervised loss has difficulty generating the adversarial example that maximizes the risk of feature misalignment. For example, the self-supervised loss forces the adversarial example to move closer toward the samples in the same class, as shown in the blue line marked with 'X' in Figure~\ref{fig:overview:inner}. This adversarial example is not the worst-case because it cannot widen the cluster of its class (i.e., maximize the radius of the manifold of the class). In the outer step, the self-supervised loss separates even samples in the same class, as shown in the red line marked with 'X' in Figure~\ref{fig:overview:outer}, which may hinder the clustering of the class.

\smallskip\noindent\textbf{Effective positive and negative mining.}
The contrastive view $A$ of the adversarial example $\tilde{x}_i$ as an anchor is filled with other anchors from the batch in previous adversarial contrastive learning methods, meaning that samples in the contrastive view are adversarial. Therefore, previous methods cannot guide a direction toward the correct class manifold for the adversarial anchor. Adversarial feature alignment requires effective positive and negative mining of benign samples to provide the right direction toward the correct class manifold.

\subsection{AFA Loss Function} \label{sec:afa:loss}

We design a new loss function for AFA the principles in Section~\ref{sec:afa:requirement}. AFA assumes an adversary who tries to push apart the feature of an anchor from its positives (i.e., samples in the same class) and moves the feature of the anchor towards its negatives. The worst-case adversarial example in AFA settles into the center of the manifold of a different class different from the example. The adversary's objective corresponds with maximizing the AFA loss function $L_{AFA}$ that can be formulated as follows:

\begin{equation}
\label{eq:afaloss}
    \mathcal{L}_{AFA} = \sum_{i=1}^{kn}\frac{-1}{|P_{AFA}(\tilde{x}_i')|} \sum_{\tilde{x}_p\in P_{AFA}(\tilde{x}_i')} \log\frac{e^{z_{\tilde{x}_i'}\cdot z_{\tilde{x}_p}/\tau}}{\underset{a\in A_{AFA}(\tilde{x}_i')}\sum_{} e^{z_{\tilde{x}_i'}\cdot z_a/\tau}}.
\end{equation}

$L_{AFA}$ in Eq.~\ref{eq:afaloss} is different from $L_{sup}$ in Eq.~\ref{eq:supcon} in various views. The anchor view is $\tilde{X}'=\tilde{X}+\delta$, such that an anchor $\tilde{x}_i'=\tilde{x}_i+\delta_i$ in $\tilde{X}'$ is an adversarial view of $\tilde{x}_i$ in $\tilde{X}$ with the worst-case perturbation $\delta_i$.
The contrastive view is $A_{AFA}(\tilde{x}_i'):=\tilde{X}$. In other words, positive and negative samples are selected from benign samples. Any other anchor than $\tilde{x}_i'$ is not included in the contrastive view $A_{AFA}(\tilde{x}_i')$ in $L_{AFA}$, while each anchor constructs a contrastive view in $L_{sup}$.
The positive samples in AFA loss is $P_{AFA}(\tilde{x}_i'):=\{\tilde{x}_p\in A_{AFA}(\tilde{x}_i) : \tilde{y}_p=\tilde{y}_i\}$.

Our loss function 1) is compatible to adversarial training, 2) is based on the supervised contrastive learning, and 3) provides a direction toward correct data manifold of classes. 
AFA loss function has the effectiveness of hard positive and hard negative mining because the feature of an adversarial example shifts along the manifold of benign samples whose features are easier to align.
Further, we have a different perspective from self-supervised adversarial contrastive learning methods~\cite{kim2020adversarial,jiang2020robust,ho2020contrastive,fan2021does}: we recruit all samples with the same class as the anchor from the clean multiview batch for positives.

\subsection{Training Strategy of AFA} \label{sec:afa:strategy}

\smallskip\noindent\textbf{Baseline optimization.}
The optimization algorithm of AFA is similar to TRADES~\cite{zhang2019theoretically} in that both simultaneously minimize natural and adversarial risks during the outer optimization. The optimization problem of TRADES is as follows:

\begin{equation}
\label{eq:trades}
    \text{\textbf{TRADES}} := 
    \underset{f}{\text{min}}\ \mathbb{E}
    \biggl\{
    \mathcal{L}_{CE} (f(X)Y) + 
    \underset{X^{'}\in \mathbb{B}(X,\epsilon)}{\text{max}} \beta\cdot\mathcal{L}_{KL}(f(X)f(X^{'}))
    \biggl\}.
\end{equation}

$\mathcal{L}_{KL}$ in Eq.~\ref{eq:trades} is KL divergence loss, and $\beta$ is a coefficient to regularize the importance of adversarial loss. $X^{'}$ is a set of adversarial examples that maximizes the inner loss function. Based on the AFA loss function of Eq.~\ref{eq:afaloss} and Eq.~\ref{eq:trades}, we define the baseline optimization algorithm of AFA as follows:

\begin{equation}
\label{eq:afa}
    \text{\textbf{AFA}} := 
    \underset{\theta_g}{\min}\ \mathbb{E}_{(x, y)\sim \mathcal{D}}\
    \{
    \lambda_1\mathcal{L}_{sup} + \underset{||\delta||\leq\epsilon}{\max} \lambda_2\mathcal{L}_{AFA}
    \}.
\end{equation}

AFA performs the inner optimization in Eq.~\ref{eq:afa} by maximizing the loss $\mathcal{L}_{AFA}$ in Eq.~\ref{eq:afaloss} through the projected gradient descent~\cite{madry2017towards}. This process is identical to finding the worst-case sample farthest from its true class and closest to its other classes, given the current distribution of feature space (see Figure~\ref{fig:overview:inner}).
While minimizing the AFA loss, the feature vector of the worst-case sample is relocated to the vicinity of clean samples in its true class (see Figure~\ref{fig:overview:outer}).
AFA embodies Theorem~\ref{theo:train_alignment} in action, as it concurrently conducts clustering and separation throughout training.
We believe that minimizing vanilla supervised contrastive loss $L_{sup}$ is helpful because feature values of well-aligned training samples are a reliable label for the AFA loss where ground-truth labels are not given in the feature space.

The coefficient $\lambda$ regularizes the influences of $L_{sup}$ and $L_{AFA}$. Eq.~\ref{eq:afa} is identical to the vanilla supervised contrastive learning~\cite{khosla2020supervised} when $\lambda_2=0$. On the other hand, Eq.~\ref{eq:afa} targets only the AFA loss when $\lambda_1=0$. Through the ablation study in Appendix~\ref{sup:experiments:ablation} and Table~\ref{tab:evaluation:lambda}, we set $\lambda_1=1$ and $\lambda_2=2$ for the baseline AFA. Based on the result of Table~\ref{tab:evaluation:viewsize}, we apply the 3-fold augmentation to the baseline AFA: for each original image, two derivatives are randomly transformed, and the other is preserved.

\smallskip\noindent\textbf{Pre-training with AFA.}
The most basic training method with AFA, similar to other contrastive learning techniques, involves pre-training followed by fine-tuning. In this case, AFA optimization denoted by Eq.~\ref{eq:afa} is used solely to pre-train a neural network's feature extractor. Subsequently, traditional adversarial training methods that utilize the loss function of the output layer are employed to fine-tune either the neural network's linear classifier or its entire parameters.

\smallskip\noindent\textbf{Joint optimization method.}
AFA optimization impacts only the feature extractor, not the linear classifier. Therefore, it is possible to use AFA in conjunction with other adversarial training methods. Algorithm~\autoref{alg:joint} describes the procedure of our joint optimization that combines AFA with TRADES. We use original inputs in a training batch to optimize TRADES loss and augmented inputs to optimize AFA loss. In the inner step, we individually generate adversarial examples that maximize the AFA loss denoted by Eq.\ref{eq:afaloss} and the KL loss from TRADES. In the outer step, we use these adversarial examples to minimize both Eq.\ref{eq:afa} and Eq.~\ref{eq:trades} simultaneously.
Due to the considerably higher AFA loss values compared to TRADES loss values, we use the TRADES loss as is to update the network parameter but regularize the AFA loss value by a factor of 0.1.
This approach can also be combined with other training methods, not just TRADES.

\begin{algorithm*}[t]
	\caption{Joint optimization of AFA and TRADES}
	\label{alg:joint}
        \begin{algorithmic}[1]
        \State \textbf{Input:} Neural network $f$ that consists of feature extractor $g$ and linear classifier $h$
        \State \textbf{Input:} Network parameters $\theta_f, \theta_g, \theta_h$
        \State \textbf{Input:} Training batch $\tilde{X}$, class label $Y$, perturbation size $\epsilon$
        \State \textbf{Output:} Updated network parameters
        
        \State $\tilde{X}_{TRADES}, \tilde{X}_{AFA}$ $\gets$ Separate($\tilde{X}$)
        \Comment{Separate the batch into original inputs and augmented inputs.} 
        
        \State \textbf{// Inner maximization phase}
        \State $\delta_{AFA}, \delta_{TRADES}$ $\gets$ Uniform($-\epsilon, \epsilon$)
        \Comment{Initialize adversarial perturbations.}
        \State $\delta_{AFA}$ $\gets$ $\underset{\delta_{AFA}}{\text{max}} \mathcal{L}_{AFA}(\theta_g, \tilde{X}_{AFA}, Y, \delta_{AFA})$
        \Comment{Derive adversarial perturbation maximizing AFA loss for the augmented inputs.}
        \State $\delta_{TRADES}$ $\gets$ $\underset{\delta_{TRADES}}{\text{max}} \mathcal{L}_{KL}(\theta_f, \tilde{X}_{TRADES}, \delta_{TRADES})$
        \Comment{Derive adversarial perturbation maximizing TRADES loss for the original inputs.}
        
        \State \textbf{// Outer minimization phase}
        \State $\mathcal{L}_{AFA}$ $\gets$ $\lambda_1\mathcal{L}_{sup}(\tilde{X}_{AFA}, Y) + \lambda_2\mathcal{L}_{AFA}(\tilde{X}_{AFA}, Y, \delta_{AFA})$
        \Comment{Measure AFA training loss.}
        \State $\mathcal{L}_{TRADES}$ $\gets$ $\mathcal{L}_{CE} (f(\tilde{X}_{TRADES})Y) + \beta\cdot\mathcal{L}_{KL}(f(\tilde{X}_{TRADES})f(\tilde{X}_{TRADES}+\delta_{TRADES}))$
        \Comment{Measure TRADES training loss.}
        \State $\mathcal{L}_{joint}$ $\gets$ $\mathcal{L}_{TRADES} + 0.1 \cdot \mathcal{L}_{AFA}$
        \Comment{Derive the final training loss}
        \State $\theta_f$ $\gets$ Update($\theta_f$, $\mathcal{L}_{joint}$)
        \Comment{Update parameters with the joint loss}
        \end{algorithmic}
\end{algorithm*} 
\section{Evaluation} \label{sec:evaluation}

\begin{table}[t]
\begin{footnotesize}
\begin{center}
\caption{Clean and robust accuracy of different training methods for ResNet-18. The best results among adversarial training methods are highlighted in bold. The attack parameters for PGD are the same as Table~\ref{tab:alignment:feature_space:cifar10}. AA denotes AutoAttack~\cite{croce2020reliable} which ensembles multiple parameter-free attacks for more reliable robustness verification.
}
\label{tab:evaluation:accuracy}
\renewcommand{\arraystretch}{1.4}
\begin{tabular}{l||M{0.5cm}M{0.5cm}M{0.5cm}|M{0.5cm}M{0.5cm}M{0.5cm}}
\toprule
\multicolumn{1}{c||}{\multirow{2}{*}{\textbf{Training Method}}} & \multicolumn{3}{c|}{\textbf{CIFAR10 (\%)}} & \multicolumn{3}{c}{\textbf{CIFAR100 (\%)}}\\
& \textbf{Clean} & \textbf{PGD} & \textbf{AA} & \textbf{Clean} & \textbf{PGD} & \textbf{AA}\\ \midrule[0.2ex]
\rowcolor{gray!10}\multicolumn{7}{c}{Natural Training Methods} \\
Cross-entropy
& 92.87 &     0 &     0 & 75.05 &     0 &  0.02 \\
SupCon~\cite{khosla2020supervised}
& 93.44 & 17.90 &  0.02 & 65.99 & 0.02 &  0.08 \\
\hline
\rowcolor{gray!10}\multicolumn{7}{c}{Adversarial Training Methods} \\
PGD AT~\cite{madry2017towards}
& 83.77 & 49.18 & 38.15 & 57.69 & 25.78 & 14.57 \\
TRADES $\beta$=$1$~\cite{zhang2019theoretically}
& 86.49 & 46.29 & 34.70 & 62.61 & 21.34 & 13.64 \\
TRADES $\beta$=$6$~\cite{zhang2019theoretically}
& 80.84 & 51.23 & 40.03 & 57.04 & 28.25 & 15.58 \\
AWP~\cite{wu2020adversarial}
& 80.40 & 54.71 & 49.57 & -     & -     & -     \\
\hline
\rowcolor{gray!10}\multicolumn{7}{c}{Adversarial Contrastive Learning Methods} \\
AdvCL~\cite{fan2021does}
& 80.24 & 53.93 & 41.08 &  58.4 & 29.86 & 16.83 \\
~~+ A-InfoNCE~\cite{yu2022adversarial}
& 83.78 & 54.36 & 41.32 & 59.16 & \textbf{30.47} & 17.23 \\
\rowcolor{gray!20}\textbf{AFA (ours)}
& \textbf{91.01} & \textbf{57.77} & \textbf{52.05} & \textbf{66.14} & 29.97 & \textbf{19.02} \\
\hline
\bottomrule
\end{tabular}
\end{center}
\end{footnotesize}
\end{table}

In this section, we evaluate the performance of our method, Adversarial Feature Alignment (AFA), from various perspectives.
We evaluate the baseline performance of AFA as an adversarial contrastive learning method, and verify our improvements in accuracy to clean samples and adversarial examples (\S\ref{sec:evaluation:baseline}).
We observe how well aligned the feature space learned by our robust training method (\S\ref{sec:evaluation:feature}).
In the ablation study, we evaluate the performance change of adversarial feature alignment concerning training settings (\S\ref{sec:evaluation:ablation}).
We apply the joint approach of AFA to the state-of-the-art adversarial training method and benchmark the performance of AFA (\S\ref{sec:evaluation:benchmark}).
The experimental settings for AFA and other training methods are described in the Appendix~\ref{sup:settings}.

\subsection{Baseline Performance of AFA} 
\label{sec:evaluation:baseline}

\smallskip\noindent\textbf{Tradeoff between robustness and accuracy.}
Table~\ref{tab:evaluation:accuracy} describes the accuracy of training methods to clean samples and adversarial examples.
We consider cross-entropy and vanilla supervised contrastive learning (SupCon)~\cite{khosla2020supervised} for natural training methods. For adversarial training methods, we consider PGD AT~\cite{madry2017towards}, TRADES~\cite{zhang2019theoretically}, and adversarial weight perturbation (AWP)~\cite{wu2020adversarial}. We also consider AdvCL~\cite{fan2021does} and A-InfoNCE~\cite{yu2022adversarial}, the state-of-the-art adversarial contrastive learning schemes.

Natural training methods achieved the highest accuracy to clean samples but misclassified most adversarial examples for both datasets.
Adversarial training methods improved robust accuracy compared to natural training, but they underwent a drop in clean accuracy.
AdvCL and A-InfoNCE yielded higher robust accuracy than PGD and TRADES, demonstrating the effectiveness of pre-training with adversarial contrastive learning. AWP, the more recent work than TRADES, aims at robustness generalization and showed the highest robust accuracy among previous methods. However, 
none of the adversarial training methods showed the best accuracies for both clean samples and adversarial examples.

AFA resulted in the best performance among all adversarial training methods in CIFAR10 and CIFAR100.
The robust accuracy of AFA on CIFAR10 was the highest for all adversarial attacks. The robust accuracy of AFA to PGD on CIFAR100 was slightly lower than A-InfoNCE. However, AFA showed the highest accuracies for AutoAttack, a parameter-free attack, implying that our methods are robust against unlearned types of attacks.
We also see that AFA's clean accuracy significantly improved from previous adversarial training methods. We cost only a 2.43\%p drop in clean accuracy compared to vanilla SupCon, and our clean accuracy was even 7.23\%p and 4.52\%p higher than A-InfoNCE and TRADES $\beta=1$ on CIFAR10, respectively. Furthermore, the clean accuracy of AFA was only 8.91\%p lower than the cross entropy and even 0.15\%p higher than the vanilla SupCon on CIFAR100. We identified from this result that AFA improves the tradeoff between robustness and accuracy.

\begin{figure}
  \centering
  \subfigure[Attack Iteration]{
    \includegraphics[width = 0.225\textwidth]{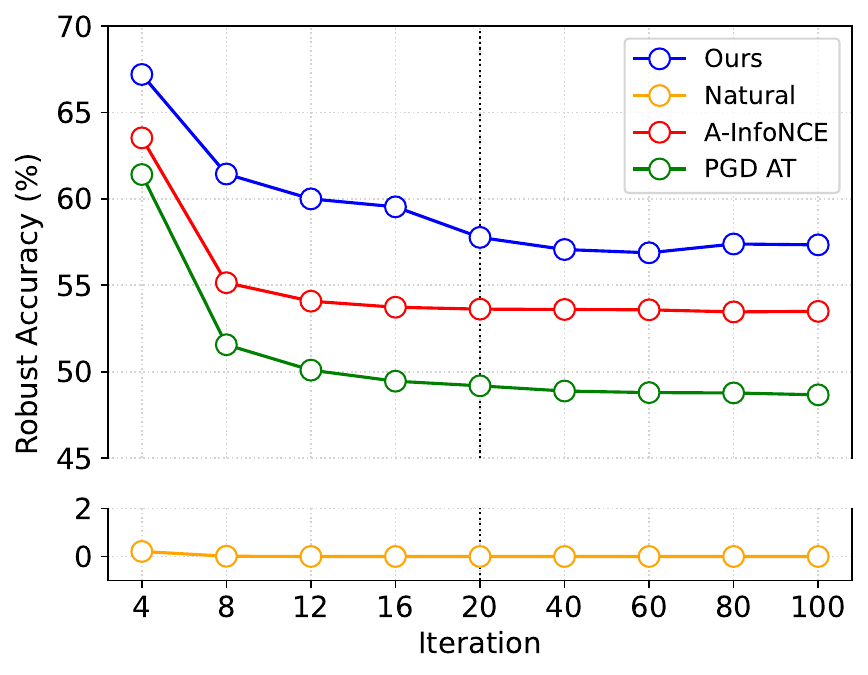}
    \label{fig:sup:evaluation:strong:iteration}
  }
  \subfigure[Perturbation Size]{
    \includegraphics[width = 0.225\textwidth]{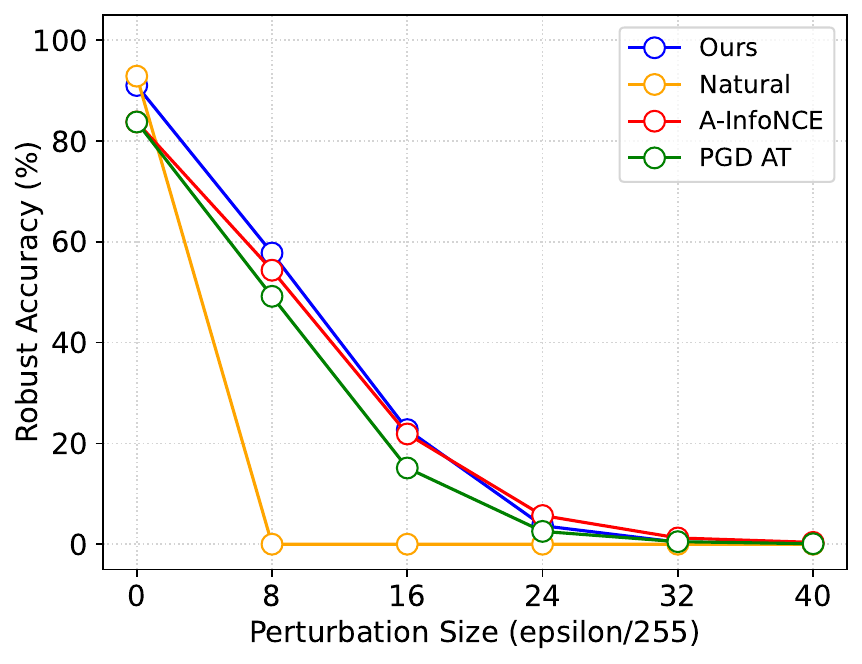}
    \label{fig:sup:evaluation:strong:persize}
  }
  \caption{Robust accuracy of training methods against varying attack strength. (a) We set $\epsilon=8/255$ and $\alpha=2/255$ for the PGD attack while changing the number of attack iterations $N$. (b) We set the number of iterations $N=20$ and $\alpha=\epsilon/4$ while changing the perturbation size $\epsilon$. The target model is ResNet-18 and experimented dataset is CIFAR10.}
  \label{fig:evaluation:strong} 
\end{figure}

\noindent\textbf{Robustness against stronger attack.}
To evaluate the performance of AFA against a more powerful attack, we consider two cases: a PGD adversary who changes the number of attack iterations with fixed perturbation size and another PGD adversary who changes perturbation size with fixed attack iterations.
In Figure~\ref{fig:sup:evaluation:strong:iteration}, the increase of attack iteration marginally reduced the robust accuracy for other methods. Our method showed the best performance than other methods for all attack iterations and even improved the accuracy against more attack iterations.
In Figure~\ref{fig:sup:evaluation:strong:persize}, the increase of perturbation size resulted in a meaningful impact on the robustness. All training methods including our method failed to avoid the reduction in the robust accuracy. AFA exhibited the best performance within the permissible perturbation size, but showed slightly lower robust accuracy than A-InfoNCE when the perturbation was sufficiently large. This result leaves future work for AFA in resolving feature misalignment that occurs with larger distortions in the input space.

\begin{table}
\begin{footnotesize}
\begin{center}
\caption{Accuracy (\%) of different training methods on additional scenarios. We could not evaluate A-InfoNCE in restricted ImageNet because of its complexity: it took more than few weeks for pretraining in our settings.}
\label{tab:evaluation:additional}
\renewcommand{\arraystretch}{1.4}
\begin{tabular}{l||M{1.0cm}M{1.0cm}|M{1.0cm}M{1.0cm}}
\toprule
 & \multicolumn{2}{c}{\textbf{CIFAR100}} & \multicolumn{2}{c}{\textbf{Restricted ImageNet}} \\
 & \multicolumn{2}{c}{\textbf{ResNet-50}} & \multicolumn{2}{c}{\textbf{ResNet-18}} \\ \cmidrule{2-5}
\textbf{Method} & \textbf{Clean} & \textbf{PGD} & \textbf{Clean} & \textbf{PGD} \\ \midrule[0.2ex]
TRADES $\beta$=1~\cite{zhang2019theoretically}
& 61.76 & 23.08 & 90.34 & 83.04 \\
TRADES $\beta$=6~\cite{zhang2019theoretically}
& 57.17 & 28.92 & 89.56 & 84.27 \\
A-InfoNCE~\cite{yu2022adversarial}
& 60.40 & 31.47 & - & - \\
\rowcolor{gray!20}\textbf{AFA (Ours)}
& \textbf{68.50} & \textbf{41.28} & \textbf{90.55} & \textbf{84.79} \\
\bottomrule
\end{tabular}
\end{center}
\end{footnotesize}
\end{table}

\smallskip\noindent\textbf{Performance of AFA on different scenarios.}
To evaluate the generalizability of our method, we conducted experiments on additional scenarios. We verified TRADES, A-InfoNCE, and AFA for ResNet-50, a more extensive model than ResNet-18, on CIFAR100.
We also evaluated these methods on restricted ImageNet, the larger dataset with more samples and higher resolution.
We aimed to validate a dataset distinct from CIFAR10 and CIFAR100. Due to the computational demands of the original ImageNet, we opted for the widely-used restricted ImageNet as a substitute.
As seen in the left two columns of Table~\ref{tab:evaluation:additional}, all methods improved their CIFAR100 accuracies from those on ResNet-18. Nevertheless, AFA achieved remarkable improvements compared to other training methods and still demonstrates the highest accuracies. The robust accuracy of AFA against PGD was 9.81\%p higher than that of A-InfoNCE on CIFAR100 with ResNet-50. This indicates the beneficial impact of model capacity on the feature alignment task in complex datasets. In the right two columns of the Table, when using the ResNet-18 model on the Restricted ImageNet dataset, AFA showed better performance in terms of both robustness and accuracy than TRADES, implying its generalizability.

\subsection{Feature Alignment and Generalization} \label{sec:evaluation:feature}

\begin{figure*}[t]
  \centering
  \subfigure[PGD Training~\cite{madry2017towards}]{
    \includegraphics[width = 0.3\textwidth]{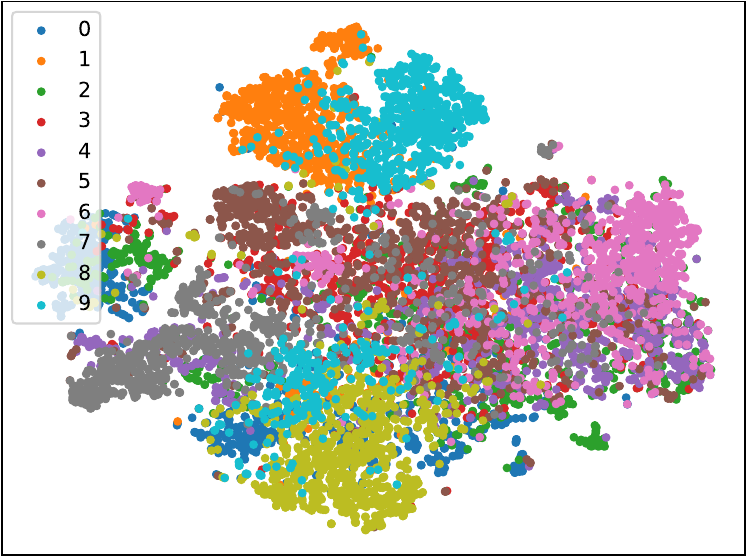}
    \label{fig:evaluation:separation:pgd}
  }
  \subfigure[AdvCL~\cite{fan2021does}]{
    \includegraphics[width = 0.3\textwidth]{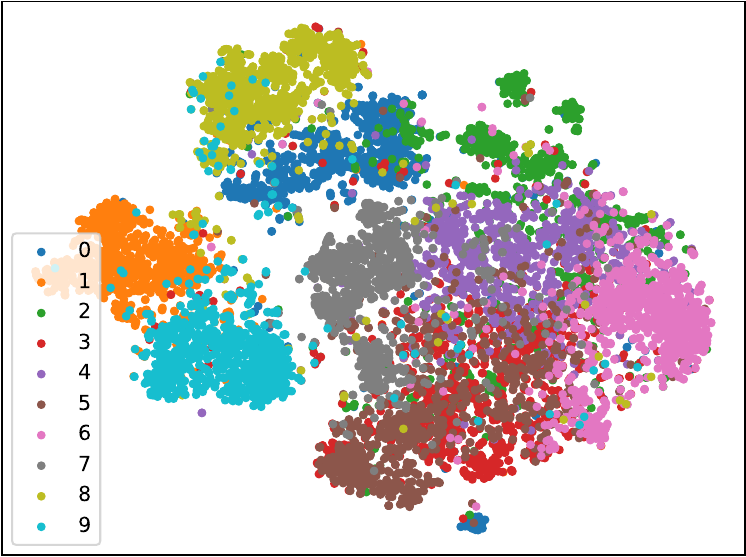}
    \label{fig:evaluation:separation:advcl}
  }
  \subfigure[Adversarial Feature Alignment (Ours)]{
    \includegraphics[width = 0.3\textwidth]{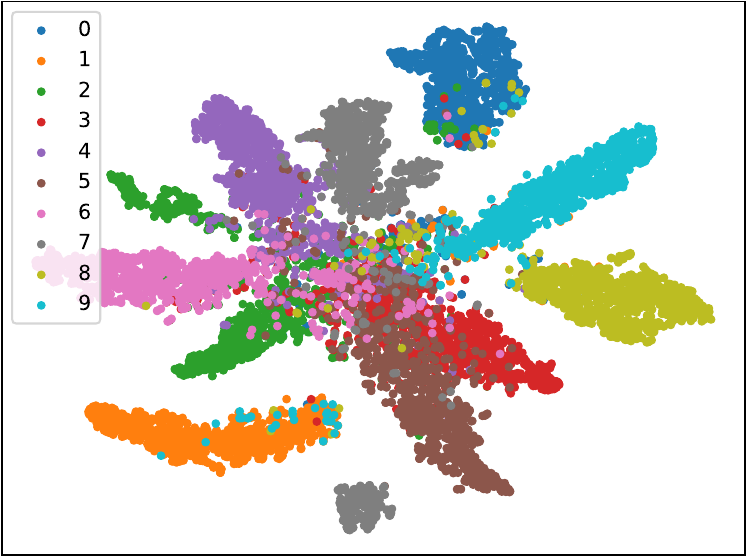}
    \label{fig:evaluation:separation:scal}
  }
  
  \caption{t-SNE visualization of feature spaces represented by adversarial training methods on CIFAR10. While PGD and AdvCL result in widespread feature spaces, our adversarial feature alignment method separates classes more distinctly.}
  \label{fig:evaluation:separation}
\end{figure*}

\textbf{Visualization of feature space.}
Figure~\ref{fig:evaluation:separation} describes that AFA, in a supervised manner, led to a remarkable alignment in the feature topology with the same network capacity. We identified that the feature space was not clearly separated and clustered by other methods. Although some samples overlap in the center for AFA, the remaining ones are more clearly separated from other classes. We chose CIFAR10 as a representative result due to its balanced number of samples and classes. We expect that the visualization for CIFAR100, which had lower accuracy, might be less clear, while restricted ImageNet, with its higher accuracy, should produce a distinct feature space similar to CIFAR10.

\begin{table}
\begin{footnotesize}
\begin{center}
\caption{Analysis of separation and alignment in the penultimate layer output of ResNet-18 models, trained by various methods, on CIFAR10. 'Clean' column compares training with clean test samples, while 'PGD' column contrasts training samples with PGD test examples, using the same attack parameters as in Table~\ref{tab:alignment:feature_space:cifar10}.}
\label{tab:evaluation:alignment:cifar10}
\renewcommand{\arraystretch}{1.4}
\begin{tabular}{l||M{0.7cm}M{0.7cm}|M{0.7cm}M{0.7cm}}
\toprule
 & \multicolumn{2}{c|}{\textbf{Min Separation}} & \multicolumn{2}{c}{\textbf{Max Clustering}}\\
\textbf{Training Method} & \textbf{Clean} & \textbf{PGD} & \textbf{Clean} & \textbf{PGD}\\ \midrule[0.2ex]
\rowcolor{gray!10}\multicolumn{5}{c}{Cross-entropy based Methods} \\
Cross-entropy
& 1.358 & 0.938 & 23.496 & 25.067 \\
PGD AT~\cite{madry2017towards}
& 0.970 & 1.480 & 22.659 & 24.632 \\
TRADES $\beta=1$~\cite{zhang2019theoretically}
& 1.789 & 1.608 & 21.980 & 23.352 \\
TRADES $\beta=6$~\cite{zhang2019theoretically}
& 1.516 & 1.460 & 22.743 & 23.355 \\
\rowcolor{gray!10}\multicolumn{5}{c}{Contrastive Learning Methods} \\
SupCon~\cite{khosla2020supervised}
& 3.547 & 1.447 & 19.789 & 20.919 \\
AdvCL~\cite{fan2021does}
& 2.094 & 1.755 & 19.989 & 20.261 \\
~~+ A-InfoNCE~\cite{yu2022adversarial}
& 1.790 & 1.526 & 19.758 & 20.486 \\
\rowcolor{gray!20}\textbf{AFA (Ours)}
& \textbf{3.674} & \textbf{3.260} & \textbf{18.095} & \textbf{19.021} \\
\bottomrule
\end{tabular}
\end{center}
\end{footnotesize}
\end{table}

\smallskip\noindent\textbf{Alignment in feature space.}
We evaluate whether AFA improves the alignment of feature space by increasing the separation factor and reducing the clustering factor. Table~\ref{tab:evaluation:alignment:cifar10} describes the minimum separation and maximum clustering factors of training methods. Adversarial examples worsen the alignment than clean test samples for all training methods. Contrastive learning-based methods generally showed better performance in alignment than cross-entropy-based methods: The distance between different classes was further, and the maximum radius of one class was shorter.
The separation of clean samples through the vanilla SupCon was improved remarkably, that of adversarial examples was worse than cross-entropy-based robust training methods. The reason is that the feature alignment that only targets clean samples cannot generalize features of adversarial examples. AdvCL and A-InfoNCE, which use an instance-wise loss function, also showed no significant improvement in separation. On the other hand, AFA performed best in separation and clustering for all types of samples.
We further verified the generalization performance of AFA through local Lipschitzness in Appendix~\ref{sup:experiments:lipschitz} and Table~\ref{tab:evaluation:lipschitz}. The results show that AFA demonstrates its robustness against adversarial examples that worsen local Lipschitzness.

\subsection{Ablation Study} \label{sec:evaluation:ablation}

\begin{table}[t]
\begin{footnotesize}
\begin{center}
    \caption{Accuracy comparison of different supervised contrastive learning optimization methods on CIFAR10 and CIFAR100 using ResNet-18. The first row shows results using a joint loss function without optimization of AFA loss (Eq.~\ref{eq:joint}). The second and third rows depict pre-training with Eqs.~\ref{eq:naive} and \ref{eq:afa}, followed by adversarial linear fine-tuning.}
    \label{tab:evaluation:optimization}
    \renewcommand{\arraystretch}{1.2}
    \begin{tabular}{l||M{0.9cm}M{0.9cm}|M{0.9cm}M{0.9cm}}
        \toprule
        \textbf{Optimization}   & \multicolumn{2}{c}{\textbf{CIFAR10 (\%)}} & \multicolumn{2}{c}{\textbf{CIFAR100 (\%)}} \\
        \textbf{Method}   & \textbf{Clean} & \textbf{PGD}   & \textbf{Clean} & \textbf{PGD} \\ \midrule[0.2ex]
        Eq.~\ref{eq:joint}     & 84.46 & 41.42 & 57.45 & 22.42 \\
        Eq.~\ref{eq:naive}     & 87.6 & 43.04 & 59.25 & 28.74 \\
        \rowcolor{gray!20}
        Eq.~\ref{eq:afa} (AFA, ours)           & \textbf{91.01} & \textbf{57.77} & \textbf{66.14} & \textbf{29.97} \\
        \bottomrule
    \end{tabular}
\end{center}
\end{footnotesize}
\end{table}

\smallskip\noindent\textbf{Comparison with other optimizations.}
We identify the necessity of AFA for robustness-accuracy tradeoff instead of other variations of supervised contrastive learning. We trained the neural network using two optimization strategies and compared the performance with our method in Table~\ref{tab:evaluation:optimization}.

Similar to the joint approach of AFA, one can consider a loss function that combines supervised contrastive loss and adversarial cross-entropy loss. Following \cite{bui2021understanding}, we define Eq~\ref{eq:joint}:
\begin{equation}
\label{eq:joint}
    \underset{\theta}{\min}\ \mathbb{E}_{(x, y)\sim \mathcal{D}}
    \{
    \mathcal{L}_{SCL} + \underset{||\delta||\leq\epsilon}{\max} \mathcal{L}_{CE}(x+\delta, y; \theta)
    \}.
\end{equation}
Eq.~\ref{eq:joint} first generates an adversarial example with maximized cross-entropy loss and then minimizes the joint loss function in each training epoch. The difference between AFA and Eq.~\ref{eq:joint} is the inner maximization of AFA loss. For the first row in Table~\ref{tab:evaluation:optimization}, we trained an entire network with Eq.~\ref{eq:joint} from the scratch for 100 epochs.
We can also consider another strategy that does not take hard positive and negative mining. We define Eq.~\ref{eq:naive}:
\begin{equation}
\label{eq:naive}
    \underset{\theta_g}{\min}\ \mathbb{E}_{(x, y)\sim \mathcal{D}}\
    \underset{||\delta||\leq\epsilon}{\max} \mathcal{L}_{SCL}
\end{equation}
where the min-max problem is solved with vanilla SupCon loss of only adversarial examples. For this strategy, we performed pre-training and fine-tuning with our baseline settings.

Results in Table~\ref{tab:evaluation:optimization} show that our strategy is the most accurate to clean and adversarial samples for both datasets. Eq.~\ref{eq:naive} yields better accuracies than Eq.~\ref{eq:joint}, indicating that pre-training with SupCon loss is quite effective for accuracy and robustness. We empirically found that Eq.~\ref{eq:joint} converges in the intermediate epoch. However, maximizing vanilla SupCon loss considerably decreased the accuracy compared to our strategy.

\subsection{Improving Adversarial Training via AFA}
\label{sec:evaluation:benchmark}

In this section, we combine AFA with state-of-the-art adversarial training methods to achieve enhanced performance. Recent studies have shown that data augmentation based on generated models, especially the denoising diffusion probabilistic model (DDPM)~\cite{ho2020denoising}, is helpful in adversarial training~\cite{gowal2021improving,rebuffi2021data,wang2023better}. Notably, Wang et al.~\cite{wang2023better} achieved top-1 performance on RobustBench in terms of accuracy for clean samples and AutoAttack by utilizing a more recent diffusion model, the elucidating diffusion model (EDM)~\cite{karras2022elucidating}. They composed training batches with original and generated data, setting the original-to-generated ratio at 0.3. Labels for the original and generated data were derived respectively from the dataset's ground truth and the predictions of a pre-trained non-robust model.

Since their focus was on data augmentation, they continued to use TRADES as the training loss function. Therefore, it is expected that replacing TRADES with AFA, which had outperformed TRADES in baseline experiments, would further enhance performance. We employed 1 million samples of data generated by EDM, as per the basic setup of Wang et al.~\cite{wang2023better}, for training the WideResNet-28-10 model. We adopted the joint optimization of AFA and TRADES of Section~\ref{sec:afa:strategy} with $\lambda_1=1$, $\lambda_2=5$, and $\beta=5$, respectively.

\begin{table}[t]
\begin{footnotesize}
\begin{center}
    \caption{
    Comparison with the latest adversarial training technique~\cite{wang2023better} on CIFAR10 using WideResNet-28-10. Note: $^*$ indicates training samples generated via a diffusion model.
    }
    \label{tab:evaluation:benchmark}
    \renewcommand{\arraystretch}{1.2}
    \begin{tabular}{l||M{1.1cm}M{0.7cm}M{0.7cm}|M{0.7cm}M{0.7cm}M{0.7cm}}
        \toprule
        \textbf{}   & \multicolumn{3}{c|}{\textbf{Settings}} & \multicolumn{3}{c}{\textbf{Accuracy (\%)}} \\
        \textbf{Method}   & \textbf{Generated$^*$} & \textbf{Batch} & \textbf{Epoch} & \textbf{Clean}   & \textbf{PGD} & \textbf{AA} \\ \midrule[0.2ex]
        \multirow{2}{1.3cm}{Wang et al. ~\cite{wang2023better}}
                        &  1M &  512 &  400 & 90.91  & 64.18 & 63.07  \\
                        &  1M & 1024 &  800 & 91.18  & 65.12 & 62.39  \\
        \rowcolor{gray!20}
        Joint AFA       &  1M &  512 &  400 & 91.79  & 64.31 & 62.94  \\
        \bottomrule
    \end{tabular}
\end{center}
\end{footnotesize}
\end{table}

Table~\ref{tab:evaluation:benchmark} presents the performance comparison between the pure approach of Wang et al.~\cite{wang2023better} and the approach that additionally combines AFA. For the same batch size and training epochs, joint AFA demonstrated slightly lower accuracy against AutoAttack compared to Wang et al.~\cite{wang2023better}, but it showed higher accuracy against PGD adversarial examples. Moreover, joint AFA improved the accuracy on clean samples by 0.88 percentage points over Wang et al.
Under the same conditions, AFA exhibited higher accuracy than Wang et al.~\cite{wang2023better}, who applied a batch size of 1024 and training epochs of 800, for both clean samples and AutoAttack.

Wang et al. also conducted experiments on a larger model, the WRN-70-16 with 267M parameters, in comparison to the WRN-28-10, which has 36M parameters. Although our performance on the WRN-28-10 was somewhat lower than their results on the WRN-70-16, we demonstrated comparable performance with the same model size, suggesting potential generalization to that model. Due to our limited computing resources, further experiments with more generated data and model parameters are left for future work.
\section{Discussion} \label{sec:discussion}

\smallskip\noindent\textbf{Implication of our work.}
The prevailing understanding is that a Lipschitz-bounded linear classifier, due to its large margin, yields better generalization performance~\cite{von2004distance,xu2009robustness}. Previously, it was believed that sufficient separability in the dataset's input space is key for model generalization~\cite{yang2020closer}. However, our research challenges this notion by showing that even with distinguishable classes, the distance metric in the input space can hinder generalization due to greater distances among same-class samples. We advocate for the generalization of the 'real' linear classifier in DNNs through the alignment of the feature space, as opposed to the static input space. Our findings indicate that low accuracy of a 1-NN classifier in the feature space suggests considerable overlap in feature distributions between classes, compromising the margin. This misalignment can lead to a reduced margin in the DNN's logit layer, increasing misclassifications. Adversarial Feature Alignment (AFA), by directly targeting feature misalignment, can enhance the margin for the linear classifier more effectively than other methods, thereby improving classification generalization.

\smallskip\noindent\textbf{Suitable fine-tuning method of AFA pretraining.}
Contrastive learning fine-tuning can be broadly classified into four methods: Standard Linear Fine-tuning (SLF), Adversarial Linear Fine-tuning (ALF), Standard Full Fine-tuning (SFF), and Adversarial Full Fine-tuning (AFF). Standard methods focus on minimizing loss on clean samples without generating adversarial examples during training, whereas adversarial methods create and minimize loss on adversarial examples. Linear methods only update the linear classifier's parameters, keeping the feature extractor constant, while full methods update all DNN layer parameters. In this paper, we apply ALF to AFA, unlike other adversarial contrastive learning methods that use AFF. We found that AFF, a full fine-tuning method, can degrade the feature extractor. Although the model performance using AFF after AFA pretraining is similar to existing methods, our primary interest lies in pretraining the feature extractor. Developing a novel fine-tuning method suited to AFA is left for future research.

\smallskip\noindent\textbf{Adaptive attacker to our approach.}
An adaptive attacker, armed with white-box knowledge of the model, dataset, and training method, aims to compromise AFA by manipulating input images to misalign class features, causing misclassification. This strategy involves using gradients that maximize AFA loss, mirroring AFA's internal optimization process. We tested this attack on a model trained with baseline AFA using 20-PGD iterations and noted a success rate of 28.99\%, lower than the 42.23\% achieved by the original PGD. These results indicate that AFA maintains robustness against such adaptive attacks.
\section{Related Work} \label{sec:related_work}

\smallskip\noindent\textbf{Adversarial defense and robust training.}
Adversarial training~\cite{goodfellow2014explaining,bai2021recent} is a method used alongside techniques like adversarial detection~\cite{grosse2017statistical, freitas2020unmask, huang2019model, li2017adversarial, ma2018characterizing, lee2018simple, liu2019detection, lu2017safetynet} and denoising~\cite{liao2018defense, samangouei2018defense, meng2017magnet, das2018shield, cho2020dapas, guo2018countering, abusnaina2021latent} as one of the most effective methods for enhancing neural network robustness against adversarial examples. The core idea of adversarial training is to learn from adversarial examples created with worst-case perturbations~\cite{goodfellow2014explaining}. This approach is conceptualized as a saddle point problem~\cite{madry2017towards}, where the inner maximization and outer minimization problems are jointly optimized for the cross-entropy loss, with methods like projected gradient descent used to maximize this loss~\cite{madry2017towards}. The following works have built upon this foundation~\cite{madry2017towards}, addressing issues like computational complexity~\cite{shafahi2019adversarial,wong2020fast} and overfitting~\cite{rice2020overfitting}.

\smallskip\noindent\textbf{Robustness-accuracy tradeoff.}
Several studies have highlighted an inherent tradeoff between robustness and accuracy in neural networks~\cite{tsipras2018robustness,zhang2019theoretically,raghunathan2019adversarial}. This tradeoff is partly attributed to the conflicting goals of natural and robustness-focused training. Robust generalization~\cite{rade2021helper,raghunathan2020understanding,chen2020more,wu2021wider,wu2020adversarial}  has become a key area of focus to address this issue. For instance, a network's robustness can be certified over a radius $\epsilon$ by estimating its Lipschitz bounds~\cite{yang2020closer,lee2020lipschitz,leino2021globally}. While Yang et al.~\cite{yang2020closer} suggested that a low local Lipschitz bounded classifier can resolve the tradeoff in separated datasets, our research in Section~\ref{sec:alignment} indicates that an accurate classifier requires not just separation, but also alignment within the dataset. Our empirical findings demonstrate that datasets which are separated but not aligned are prone to misclassifications.

To address the robustness-accuracy tradeoff, various strategies have been proposed. TRADES~\cite{zhang2019theoretically}  regularizes both standard and robust errors. Wang et al.~\cite{wang2020improving}  developed a surrogate loss function facilitating misclassification-aware regularization. Raghunathan et al.'s robust self-training leverages additional unlabeled data~\cite{raghunathan2020understanding}. Zhang et al.'s approach adjusts the loss function weight based on the distance from the decision boundary~\cite{zhang2021geometry}. Helper-based adversarial training reduces the excessive margin in decision boundaries with highly perturbed examples~\cite{rade2022reducing}. Notably, Wang et al.~\cite{wang2023better} achieved cutting-edge results in adversarial training using data augmentation with diffusion models. In Section~\ref{sec:evaluation:benchmark}, we demonstrate how the integration of AFA as a joint optimization algorithm can further enhance these approaches.

\smallskip\noindent\textbf{Adversarial contrastive learning.}
While cross-entropy as a standalone loss function has its limitations~\cite{sukhbaatar2014training,zhang2018generalized,cao2019learning}, contrastive learning has shown promising results in feature generalization~\cite{chen2020simple,zhang2021unleashing,tian2020contrastive,ho2020contrastive,hjelm2018learning,henaff2020data,khosla2020supervised}. Several studies have leveraged contrastive learning to enhance adversarial robustness~\cite{kim2020adversarial,jiang2020robust,fan2021does,yu2022adversarial}. Typically, these works focus on robust pre-training for instance discrimination, maximizing self-supervised contrastive losses without using class labels, which might not sufficiently align the feature space. Although pseudo labels from ClusterFit~\cite{yan2020clusterfit} are used in AdvCL~\cite{fan2021does} for pre-training, they are limited to cross-entropy loss. A-InfoNCE~\cite{yu2022adversarial} improved upon this with distinct positive and negative mining strategies. Our work explores the supervised contrastive approach~\cite{khosla2020supervised} for robust pre-training, aligning feature representations more effectively than self-supervised methods. Unlike previous studies that used supervised contrastive learning only in the outer minimization step~\cite{bui2021understanding}, our method prioritizes effective feature space separation and clustering through tailored loss functions and optimization techniques.
\section{Conclusion} \label{sec:conclusion}

Adversarial training is crucial for securing deep neural networks, yet its application is often limited by the tradeoff between robustness and accuracy in real-world scenarios. Our research challenges existing beliefs about data distribution and alignment, revealing that misalignment in seemingly separated datasets contributes to this tradeoff. We introduced 'Adversarial Feature Alignment', a robust pre-training method utilizing contrastive learning, to effectively address this issue. Our approach not only identifies but also rectifies the misclassifying tendencies of feature spaces. The results show that our method achieves enhanced robustness with minimal accuracy loss compared to existing approaches. Given that our method is based on a standalone optimization problem, we anticipate future enhancements through advanced supervised contrastive learning techniques~\cite{graf2021dissecting,cui2021parametric}.

\appendix

\section*{Appendices}

\section{Organization} \label{sup:organization}

This section summarizes the organization of the Appendices of this paper.
In Appendix~\ref{sup:proof}, we provide proofs of Lemma~\ref{lem:accuracy_alignment} and Theorem~\ref{theo:train_alignment}.
In Appendix~\ref{sup:settings}, we describe detailed experimental settings for our main experiments.
In Appendix~\ref{sup:alignment} and from Table~\ref{tab:sup:alignment:feature_space:cifar10} to Table~\ref{tab:sup:misalignment}, we further observe the correlation between the feature alignment and the prediction of a neural network. In Appendix~\ref{sup:experiments}, we verify the performance of AFA with additional experiments and perform ablation studies.
\section{Proofs} \label{sup:proof}

\subsection{Proof of Lemma~\ref{lem:accuracy_alignment}}

\begin{proof}
Assume the 1-nearest neighbor classifier $f_\text{1-nn}$ in the binary classification problem as follows:

\begin{align*}
    f_\text{1-nn}(x)=\frac{1}{2R}\cdot\left(\mathsf{dist}(x,\mathcal{X}_{\text{train}}^{(0)}), \mathsf{dist}(x,\mathcal{X}_{\text{train}}^{(1)})\right).
\end{align*}

We define $f_\text{1-nn}^{(i)}=\mathsf{dist}(x,\mathcal{X}_{\text{train}}^{(i)})/2R$. Let $y$ and $j$ be the ground-truth class and the other class of an input $x$, respectively. If the distribution is $R$-aligned, we have $\mathsf{dist}(x,\mathcal{X}_\text{train}^\text{(j)})\geq 2r$ by Definition~\ref{def:separation} and $\mathsf{dist}(x,\mathcal{X}_\text{train}^\text{(y)})\leq 2R$ by Definition~\ref{def:clustering}. Further, we have that $r>R$ by Definition~\ref{def:alignment}, and $2r=2R+\epsilon$ holds given $\epsilon>0$. Then, we obtain the following equation:
     
\begin{align*} 
\mathsf{dist}(x,\mathcal{X}_\text{train}^\text{(j)})    &\geq 2r \\
                                                        &= 2R + \epsilon \\
                                                        &> \mathsf{dist}(x,\mathcal{X}_\text{train}^\text{(y)}).
\end{align*}

The above equation makes that $\text{argmin}(f_\text{1-nn}^{(i)})=y$ holds for all input $x$. Thus, $f_\text{1-nn}$ has accuracy of 1 on the $R$-aligned data distribution.
\end{proof}

\subsection{Proof of Theorem~\ref{theo:train_alignment}}

\begin{proof}
Assume a data distribution forms the topology illustrated in Figure~\ref{fig:introduction:alignment}. Let $x$ be the worst-case input that belongs to the ground-truth class $y$ but is closest to its other class $i$. Let $x_\text{ref}^\text{(y)}$ be the farthest sample from $x$ among training samples in $\mathcal{X}_\text{train}^\text{(y)}$, and $x_\text{ref}^\text{(j)}$ be the closest sample to $x$ among training samples in $\mathcal{X}_\text{train}^\text{(j)}$. Then, we have $\mathsf{dist}(x, x_\text{ref}^\text{(y)}) = R + R'$ and $\mathsf{dist}(x, x_\text{ref}^\text{(j)}) = 2r + R - R'$. Given $2r=2R'+\epsilon$ where $\epsilon>0$, we obtain the following:

\begin{align*} 
\mathsf{dist}(x, x_\text{ref}^\text{(j)})   &= 2r + R - R' \\
                                            &= R + R' + \epsilon \\
                                            &> R + R' \\
                                            &= \mathsf{dist}(x, x_\text{ref}^\text{(y)}).
\end{align*}

Without loss of generality, all other samples in $\mathcal{X}_\text{train}^\text{(y)}$ are closer to $x$ than $x_\text{ref}^\text{(y)}$, and all other samples in $\mathcal{X}_\text{train}^\text{(j)}$ are further from $x$ than $x_\text{ref}^\text{(j)}$. That is, $\mathsf{dist}(x,\mathcal{X}_{\text{train}}^{(j)}) > \mathsf{dist}(x,\mathcal{X}_{\text{train}}^{(y)})$ holds for all input $x$. Thus, if the data distribution is $r$-separated with $r>R'$, $\text{argmin}(f_\text{1-nn}^{(i)})=y$ satisfies for all $x$, and $f_\text{1-nn}$ has accuracy of 1.
\end{proof}

\begin{table*}[t]
\begin{footnotesize}
\begin{center}
\caption{Separation and clustering factors and the accuracy of the 1-nearest neighbor $f_\text{1-nn}$ for the input space of various datasets. The separation factor $2r$ is the minimum/average/maximum distance between two different classes. The clustering factor $2R$ is the minimum/average/maximum distance of samples within the same class.
Pixel values are normalized to the range [0, 1].
We report results with $l_0$, $l_1$, $l_2$, $l_\infty$ norm, respectively.
We abbreviate Restricted ImageNet dataset to ResImageNet.}
\label{tab:sup:alignment:input_space}
\renewcommand{\arraystretch}{1.4}
\begin{tabular}{l||M{0.7cm}M{0.7cm}M{0.7cm}|M{0.7cm}M{0.7cm}M{0.7cm}|M{0.7cm}M{0.7cm}|M{0.7cm}M{0.7cm}|M{2.5cm}}
\toprule
 & \multicolumn{6}{c|}{\textbf{Separation Factor}} & \multicolumn{4}{c|}{\textbf{Clustering Factor}} & \multirow{3}{*}{\textbf{Test Accuracy of $f_\text{1-nn}$}} \\
\textbf{Dataset} & \multicolumn{3}{c}{\textbf{Train-Train}} & \multicolumn{3}{c|}{\textbf{Train-Test}} & \multicolumn{2}{c}{\textbf{Train-Train}} & \multicolumn{2}{c|}{\textbf{Train-Test}} & \\
 & \textbf{Min} & \textbf{Avg} & \textbf{Max} & \textbf{Min} & \textbf{Avg} & \textbf{Max} & \textbf{Min} & \textbf{Max} & \textbf{Min} & \textbf{Max} & \textbf{(\%)}\\ \midrule[0.2ex]
\rowcolor{gray!20} \multicolumn{12}{c}{\textbf{Distance metric = $L_0$ norm}}\\
MNIST           &  44.0 &  85.6 & 109.0 &  47.0 & 97.49 & 128.0 & 316.0 & 369.0 & 295.0 & 393.0 & 82.23 \\
CIFAR10         &   816 &  1725 &  2099 &   840 &  1878 &  2353 &  3072 &  3072 &  3072 &  3072 & 27.06 \\
CIFAR100        &     0 &  2165 &  2968 &     0 &  2400 &  3015 &  3072 &  3072 &  3072 &  3072 & 11.23 \\
STL10           & 13387 & 20890 & 25454 & 11225 & 20825 & 25546 & 27636 & 27648 & 27639 & 27648 & 29.16 \\
\rowcolor{gray!20} \multicolumn{12}{c}{\textbf{Distance metric = $L_1$ norm}}\\
MNIST           &  14.6 & 33.08 & 47.48 & 16.34 & 38.81 & 55.19 & 220.4 & 288.2 & 204.4 & 272.1 & 96.31 \\
CIFAR10         & 65.87 & 198.0 & 291.9 & 70.09 & 220.5 & 323.3 &  2420 &  2849 &  2444 &  2845 & 39.59 \\
CIFAR100        &     0 & 255.7 & 401.0 &     0 & 295.3 & 474.9 &  1645 &  2868 &  1650 &  2841 & 19.83 \\
STL10           &  1902 &  2741 &  3595 &  1572 &  2519 &  3349 & 15671 & 23168 & 17327 & 23290 & 31.65 \\
\rowcolor{gray!20} \multicolumn{12}{c}{\textbf{Distance metric = $L_2$ norm}}\\
MNIST           & 2.398 & 4.180 & 5.502 & 2.775 & 4.688 & 6.050 & 13.94 & 16.19 & 13.32 & 15.74 & 96.92 \\
CIFAR10         & 2.830 & 5.056 & 6.961 & 2.836 & 5.475 & 7.742 & 46.20 & 51.74 & 46.48 & 51.72 & 35.46 \\
CIFAR100        &     0 & 6.224 & 10.06 &     0 & 7.095 & 11.17 & 33.13 & 52.47 & 33.75 & 51.69 & 17.56 \\
STL10           & 15.25 & 22.21 & 28.99 & 12.93 & 20.94 & 26.51 & 104.2 & 145.0 & 112.4 & 146.1 & 28.01 \\
\rowcolor{gray!20} \multicolumn{12}{c}{\textbf{Distance metric = $L_\infty$ norm}}\\
MNIST           & 0.737 & 0.927 & 0.988 & 0.812 & 0.958 & 0.988 &   1.0 &   1.0 &   1.0 &   1.0 & 78.92 \\
CIFAR10         & 0.211 & 0.309 & 0.412 & 0.220 & 0.331 & 0.443 &   1.0 &   1.0 &   1.0 &   1.0 & 18.23 \\
CIFAR100        & 0.067 & 0.371 & 0.561 & 0.114 & 0.414 & 0.604 &   1.0 &   1.0 &   1.0 &   1.0 &  4.53 \\
STL10           & 0.369 & 0.493 & 0.624 & 0.345 & 0.466 & 0.627 &   1.0 &   1.0 &   1.0 &   1.0 & 11.34 \\
ResImageNet     & 0.235 & 0.332 & 0.426 & 0.271 & 0.410 & 0.533 &   1.0 &   1.0 &   1.0 &   1.0 & 21.15 \\
\bottomrule
\end{tabular}
\end{center}
\end{footnotesize}
\end{table*}

\begin{table*}[t]
\begin{footnotesize}
\begin{center}
\caption{Separation and clustering factors for penultimate layer output of a neural network trained by different methods. The target dataset is CIFAR10. The column "Train-Test" measures the distance between training and clean test samples. The column "Train-Adv" measures the distance between training samples and PGD adversarial examples generated from clean test samples with $\epsilon=8/255$, step size $\alpha=2/255$, and attack iteration $N=20$. The distance metric is $l_1$.}
\label{tab:sup:alignment:feature_space:cifar10}
\renewcommand{\arraystretch}{1.4}
\begin{tabular}{l||M{0.7cm}M{0.7cm}|M{0.7cm}M{0.7cm}|M{0.7cm}M{0.7cm}|M{0.7cm}M{0.7cm}|M{0.7cm}M{0.7cm}M{0.7cm}M{0.7cm}}
\toprule
 & \multicolumn{4}{c|}{\textbf{Separation Factor}} & \multicolumn{4}{c|}{\textbf{Clustering Factor}} & \multicolumn{4}{c}{\textbf{Test Accuracy (\%)}} \\
& \multicolumn{2}{c}{\textbf{Train-Test}} & \multicolumn{2}{c|}{\textbf{Train-Adv}} & \multicolumn{2}{c}{\textbf{Train-Test}} & \multicolumn{2}{c|}{\textbf{Train-Adv}} & \multicolumn{2}{c}{\textbf{Clean Samples}} & \multicolumn{2}{c}{\textbf{Adv. Examples}}\\
\textbf{Method}  & \textbf{Min} & \textbf{Max} & \textbf{Min} & \textbf{Max} & \textbf{Min} & \textbf{Max} & \textbf{Min} & \textbf{Max} & \textbf{$f_\text{1-nn}$} & \textbf{$f_\text{dnn}$} & \textbf{$f_\text{1-nn}$} & \textbf{$f_\text{dnn}$}\\ \midrule[0.2ex]
\rowcolor{gray!20} \multicolumn{13}{c}{\textbf{CIFAR10, ResNet-18 result}}\\
Cross-entropy
& 1.358 & 9.030 & 0.938 & 1.836 & 20.161 & 23.496 & 23.729 & 25.067 &  92.65 &  92.87 &      0 &      0 \\
PGD AT~\cite{madry2017towards}
& 0.970 & 7.958 & 1.480 & 5.340 & 20.729 & 22.659 & 22.860 & 24.632 &  81.80 &  83.77 &  47.82 &  49.18 \\
TRADES $\beta=1$~\cite{zhang2019theoretically}
& 1.789 & 7.940 & 1.608 & 6.333 & 19.959 & 21.980 & 21.706 & 23.352 &  86.93 &  86.49 &  43.55 &  46.29 \\
TRADES $\beta=6$~\cite{zhang2019theoretically}
& 1.516 & 6.855 & 1.460 & 5.496 & 21.820 & 22.743 & 22.343 & 23.355 &  81.15 &  80.84 &  49.26 &  51.23 \\
AdvCL~\cite{fan2021does}
& 2.094 & 8.994 & 1.740 & 8.259 & 18.877 & 19.989 & 19.133 & 20.341 &  81.28 &  80.24 &  53.66 &  53.93 \\
+ A-InfoNCE~\cite{yu2022adversarial}
& 1.790 & 8.045 & 1.505 & 6.776 & 18.715 & 19.758 & 19.052 & 20.464 &  82.67 &  83.78 &  54.14 &  54.36 \\
AFA (Ours)
& 3.674 & 7.420 & 3.260 & 6.877 & 16.662 & 18.095 & 17.101 & 19.021 &  82.95 &  91.01 &  53.76 &  57.77 \\
\rowcolor{gray!20} \multicolumn{13}{c}{\textbf{CIFAR10, WideResNet-36-10 result}}\\
Cross-entropy
& 0.814 & 6.793 & 0.606 & 1.069 & 13.080 & 14.633 & 15.362 & 16.408 &  95.63 &  95.72 &  0 &  0 \\
PGD AT~\cite{madry2017towards}
& 1.341 & 6.580 & 1.223 & 3.457 & 12.501 & 14.875 & 13.757 & 14.713 &  86.60 &  86.90 &  47.10 &  47.58 \\
TRADES $\beta=1$~\cite{zhang2019theoretically}
& 1.295 & 6.417 & 1.108 & 4.082 & 14.622 & 15.303 & 15.556 & 16.581 &  88.33 &  88.60 &  46.47 &  47.11 \\
TRADES $\beta=6$~\cite{zhang2019theoretically}
& 1.423 & 5.576 & 1.225 & 4.171 & 15.407 & 16.682 & 16.494 & 17.534 &  85.07 &  85.54 &  50.06 &  51.51 \\
\bottomrule
\end{tabular}
\end{center}
\end{footnotesize}
\end{table*}

\begin{table*}[t]
\begin{footnotesize}
\begin{center}
\caption{Separation and clustering factors for penultimate layer output of a neural network trained by different methods. The target dataset is CIFAR10, and the model architecture is ResNet-18. The column "Train-Test" measures the distance between training and clean test samples. The column "Train-Adv" measures the distance between training samples and PGD adversarial examples generated from clean test samples with $\epsilon=8/255$, step size $\alpha=2/255$, and attack iteration $N=20$. The distance metrics are $l_0$, $l_1$, $l_2$, and $l_1\infty$, respectively.}
\label{tab:sup:alignment:feature_space:cifar10:metrics}
\renewcommand{\arraystretch}{1.4}
\begin{tabular}{l||M{0.7cm}M{0.7cm}|M{0.7cm}M{0.7cm}|M{0.7cm}M{0.7cm}|M{0.7cm}M{0.7cm}|M{0.7cm}M{0.7cm}M{0.7cm}M{0.7cm}}
\toprule
 & \multicolumn{4}{c|}{\textbf{Separation Factor}} & \multicolumn{4}{c|}{\textbf{Clustering Factor}} & \multicolumn{4}{c}{\textbf{Test Accuracy (\%)}} \\
& \multicolumn{2}{c}{\textbf{Train-Test}} & \multicolumn{2}{c|}{\textbf{Train-Adv}} & \multicolumn{2}{c}{\textbf{Train-Test}} & \multicolumn{2}{c|}{\textbf{Train-Adv}} & \multicolumn{2}{c}{\textbf{Clean Samples}} & \multicolumn{2}{c}{\textbf{Adv. Examples}}\\
\textbf{Method}  & \textbf{Min} & \textbf{Max} & \textbf{Min} & \textbf{Max} & \textbf{Min} & \textbf{Max} & \textbf{Min} & \textbf{Max} & \textbf{$f_\text{1-nn}$} & \textbf{$f_\text{dnn}$} & \textbf{$f_\text{1-nn}$} & \textbf{$f_\text{dnn}$}\\ \midrule[0.2ex]
\rowcolor{gray!20} \multicolumn{13}{c}{\textbf{CIFAR10, ResNet-18 results}}\\
\rowcolor{gray!10} \multicolumn{13}{c}{\textbf{Distance metric = $L_0$ norm}}\\
Cross-entropy
& 1230.0 & 1607.0 & 1047.0 & 1416.0 & 2048.0 & 2048.0 & 2046.0 & 2048.0 &  90.85 &  92.87 &      0 &      0 \\
PGD AT~\cite{madry2017towards}
&  760.0 & 1338.0 &  753.0 & 1260.0 & 2009.0 & 2042.0 & 2015.0 & 2043.0 &  76.27 &  83.77 &  46.31 &  49.18 \\
\rowcolor{gray!10} \multicolumn{13}{c}{\textbf{Distance metric = $L_1$ norm}}\\
Cross-entropy
&  1.358 &  9.030 &  0.938 &  1.836 & 20.161 & 23.496 & 23.729 & 25.067 &  92.65 &  92.87 &      0 &      0 \\
PGD AT~\cite{madry2017towards}
&  0.970 &  7.958 &  1.480 &  5.340 & 20.729 & 22.659 & 22.860 & 24.632 &  81.80 &  83.77 &  47.82 &  49.18 \\
\rowcolor{gray!10} \multicolumn{13}{c}{\textbf{Distance metric = $L_2$ norm}}\\
Cross-entropy
&  0.346 &  1.335 &  0.235 &  0.489 &  2.510 &  2.658 &  2.742 &  2.779 &  92.82 &  92.87 &      0 &      0 \\
PGD AT~\cite{madry2017towards}
&  0.186 &  1.330 &  0.336 &  1.120 &  2.608 &  2.752 &  2.742 &  2.781 &  81.85 &  83.77 &  48.73 &  49.18 \\
\rowcolor{gray!10} \multicolumn{13}{c}{\textbf{Distance metric = $L_\infty$ norm}}\\
Cross-entropy
&  0.019 &  0.110 &  0.014 &  0.026 &  0.306 &  0.417 &  0.375 &  0.430 &  92.54 &  92.87 &      0 &      0 \\
PGD AT~\cite{madry2017towards}
&  0.027 &  0.136 &  0.037 &  0.129 &  0.563 &  0.732 &  0.604 &  0.732 &  79.52 &  83.77 &  47.57 &  49.18 \\
\bottomrule
\end{tabular}
\end{center}
\end{footnotesize}
\end{table*}

\begin{table*}[t]
\begin{footnotesize}
\begin{center}
\caption{Separation and clustering factors for penultimate layer output of a neural network trained by different methods. The target dataset is CIFAR100. The column "Train-Test" measures the distance between training and clean test samples. The column "Train-Adv" measures the distance between training samples and PGD adversarial examples generated from clean test samples with $\epsilon=8/255$, step size $\alpha=2/255$, and attack iteration $N=20$. The distance metric is $l_1$.}
\label{tab:sup:alignment:feature_space:cifar100}
\renewcommand{\arraystretch}{1.4}
\begin{tabular}{l||M{0.7cm}M{0.7cm}|M{0.7cm}M{0.7cm}|M{0.7cm}M{0.7cm}|M{0.7cm}M{0.7cm}|M{0.7cm}M{0.7cm}M{0.7cm}M{0.7cm}}
\toprule
 & \multicolumn{4}{c|}{\textbf{Separation Factor}} & \multicolumn{4}{c|}{\textbf{Clustering Factor}} & \multicolumn{4}{c}{\textbf{Test Accuracy (\%)}} \\
& \multicolumn{2}{c}{\textbf{Train-Test}} & \multicolumn{2}{c|}{\textbf{Train-Adv}} & \multicolumn{2}{c}{\textbf{Train-Test}} & \multicolumn{2}{c|}{\textbf{Train-Adv}} & \multicolumn{2}{c}{\textbf{Clean Samples}} & \multicolumn{2}{c}{\textbf{Adv. Examples}}\\
\textbf{Method}  & \textbf{Min} & \textbf{Max} & \textbf{Min} & \textbf{Max} & \textbf{Min} & \textbf{Max} & \textbf{Min} & \textbf{Max} & \textbf{$f_\text{1-nn}$} & \textbf{$f_\text{dnn}$} & \textbf{$f_\text{1-nn}$} & \textbf{$f_\text{dnn}$}\\ \midrule[0.2ex]
\rowcolor{gray!20} \multicolumn{13}{c}{\textbf{CIFAR100, ResNet-18 result}}\\
Cross-entropy
& 2.622 & 14.848 & 2.187 & 13.398 & 16.082 & 21.128 & 18.795 & 21.184 &  74.53 &  75.05 &      0 &      0 \\
PGD AT~\cite{madry2017towards}
& 1.972 & 14.010 & 2.009 & 13.668 & 18.016 & 21.574 & 18.610 & 21.598 &  50.76 &  57.69 &  28.32 &  25.78 \\
TRADES $\beta=1$~\cite{zhang2019theoretically}
& 1.710 & 14.017 & 1.740 & 13.317 & 16.977 & 21.242 & 18.021 & 21.678 &  55.90 &  62.61 &  26.69 &  21.34 \\
TRADES $\beta=6$~\cite{zhang2019theoretically}
& 1.238 & 14.080 & 1.323 & 13.520 & 16.911 & 20.925 & 17.810 & 21.666 &  48.86 &  57.04 &  29.45 &  28.25 \\
AdvCL~\cite{fan2021does}
& 2.210 & 12.582 & 2.362 & 12.354 & 16.373 & 19.697 & 16.838 & 20.013 &  49.68 &  58.40 &  31.11 &  29.86 \\
+ A-InfoNCE~\cite{yu2022adversarial}
& 2.174 & 12.554 & 2.010 & 12.412 & 16.315 & 19.884 & 16.975 & 20.077 &  50.04 &  59.16 &  31.38 &  30.47 \\
AFA (Ours)
& 1.858 &  9.910 & 2.868 & 10.886 & 14.460 & 18.142 & 15.317 & 18.150 &  35.65 &  66.14 &  24.86 &  29.97 \\
\rowcolor{gray!20} \multicolumn{13}{c}{\textbf{CIFAR100, WideResNet-36-10 result}}\\
Cross-entropy
& 1.344 & 13.376 & 1.789 & 12.406 & 12.866 & 18.752 & 16.196 & 18.774 &  78.74 & 79.16 &  0.03 &  0.02 \\
PGD AT~\cite{madry2017towards}
& 0.973 & 13.716 & 1.564 & 13.384 & 15.711 & 19.419 & 16.404 & 19.619 &  58.25 &  61.54 &  26.83 &  24.97 \\
TRADES $\beta=1$~\cite{zhang2019theoretically}
& 1.737 & 13.515 & 2.247 & 13.197 & 15.439 & 19.484 & 16.447 & 19.820 &  61.34 &  64.89 &  24.90 &  23.27 \\
TRADES $\beta=6$~\cite{zhang2019theoretically}
& 1.624 & 13.950 & 1.965 & 14.121 & 17.073 & 20.243 & 17.587 & 21.093 &  54.91 &  60.47 &  29.18 &  29.39 \\
\bottomrule
\end{tabular}
\end{center}
\end{footnotesize}
\end{table*}

\begin{table*}
\begin{footnotesize}
\begin{center}
\caption{Separation and clustering factors for penultimate layer output of a neural network trained by different methods. The target dataset is Restricted ImageNet. The column "Train-Test" measures the distance between training and clean test samples. The column "Train-Adv" measures the distance between training samples and PGD adversarial examples generated from clean test samples with $\epsilon=0.005$, step size $\alpha=0.001$, and attack iteration $N=10$. The distance metric is $l_1$.}
\label{tab:sup:alignment:feature_space:res_imagenet}
\renewcommand{\arraystretch}{1.4}
\begin{tabular}{l||M{0.7cm}M{0.7cm}|M{0.7cm}M{0.7cm}|M{0.7cm}M{0.7cm}|M{0.7cm}M{0.7cm}|M{0.7cm}M{0.7cm}M{0.7cm}M{0.7cm}}
\toprule
 & \multicolumn{4}{c|}{\textbf{Separation Factor}} & \multicolumn{4}{c|}{\textbf{Clustering Factor}} & \multicolumn{4}{c}{\textbf{Test Accuracy (\%)}} \\
& \multicolumn{2}{c}{\textbf{Train-Test}} & \multicolumn{2}{c|}{\textbf{Train-Adv}} & \multicolumn{2}{c}{\textbf{Train-Test}} & \multicolumn{2}{c|}{\textbf{Train-Adv}} & \multicolumn{2}{c}{\textbf{Clean Samples}} & \multicolumn{2}{c}{\textbf{Adv. Examples}}\\
\textbf{Method}  & \textbf{Min} & \textbf{Max} & \textbf{Min} & \textbf{Max} & \textbf{Min} & \textbf{Max} & \textbf{Min} & \textbf{Max} & \textbf{$f_\text{1-nn}$} & \textbf{$f_\text{dnn}$} & \textbf{$f_\text{1-nn}$} & \textbf{$f_\text{dnn}$}\\ \midrule[0.2ex]
\rowcolor{gray!20} \multicolumn{13}{c}{\textbf{Restricted ImageNet, ResNet-18 result}}\\
Cross-entropy
& 1.644 & 17.477 & 0.888 & 12.739 & 38.466 & 46.020 & 40.671 & 50.457 &  91.28 &  91.18 &  3.72 &  4.00 \\
PGD AT~\cite{madry2017towards}
& 2.111 & 16.466 & 1.100 & 16.641 & 36.416 & 45.046 & 38.044 & 45.876 &  90.69 &  90.18 &  83.04 &  83.04 \\
TRADES $\beta=1$~\cite{zhang2019theoretically}
& 1.466 & 17.038 & 1.477 & 17.115 & 39.268 & 45.846 & 41.096 & 46.681 &  90.58 &  90.34 &  82.35 &  82.52 \\
TRADES $\beta=6$~\cite{zhang2019theoretically}
& 2.447 & 14.408 & 1.160 & 14.511 & 35.681 & 45.794 & 35.906 & 45.988 &  89.77 &  89.56 &  83.99 &  84.27 \\
\bottomrule
\end{tabular}
\end{center}
\end{footnotesize}
\end{table*}
\section{Experimental Settings} \label{sup:settings}

\subsection{Settings for Section~\ref{sec:evaluation:baseline} to  Section~\ref{sec:evaluation:ablation}} \label{sup:settings:1}

We used the SGD optimizer for all training methods. All experiments were conducted on a single NVIDIA RTX 3090 GPU.

\subsubsection{Adversarial Contrastive Learning Settings} \label{sup:settings:1:contrastive}

We describe the training details of adversarial contrastive learning methods, including AFA, AdvCL~\cite{fan2021does}, and A-InfoNCE~\cite{yu2022adversarial}.
We adhered to the configurations of prior adversarial contrastive learning research.

\begin{algorithm}[b]
	\caption{Composition of Random Transformation Functions}
	\label{alg:augmentation}
	\begin{small}
        \begin{algorithmic}[1]
        \State transform\_train = transforms.Compose([
        \State     \quad transforms.RandomResizedCrop(size=32, scale=(0.2, 1.)),
        \State     \quad transforms.RandomHorizontalFlip(),
        \State     \quad transforms.RandomApply([
        \State     \quad \quad transforms.ColorJitter(0.4, 0.4, 0.4, 0.1)
        \State     \quad ], p=0.8),
        \State     \quad transforms.RandomGrayscale(p=0.2),
        \State     \quad transforms.ToTensor(),
        \State ])
        \end{algorithmic}
	\end{small}
\end{algorithm} 

\smallskip\noindent\textbf{Pre-training.}
As a common setting,
we used a 2-layer of multi-layer perceptron that embeds the output of 128 dimensions as a projection head to pre-train the feature extractor $g$. We performed 400 pre-training epochs. We set the initial learning rate as 0.5 and 2.0 for CIFAR10 and CIFAR100, respectively, while cosine annealing is applied. The warm-up learning rate was increased from 0.01 to the initial learning rate during the first 10 epochs. We set the momentum = 0.9 and weight decay = 1e-4 for both datasets. We used a transformation algorithm listed in Algorithm~\ref{alg:augmentation} for the random augmentation. We applied the temperature $\tau=0.5$ to the contrastive losses. The considered transformation functions are random cropping, random horizontal flip, random color jittering, and random grayscale. We set the number of PGD iterations to ten for the inner maximization.

For AFA,
we augmented an original image $x$ to $\{x, \tilde{x}_1, \tilde{x}_2\}$ for the multiviewed batch, where $\tilde{x}=T(x)$ and $T$ is a random image transformation. We applied adversarial perturbations to all images in the multiviewed batch except one transformed view of each original image. That is, the adversarial multiviewed batch consists of $\{x+\delta_1, \tilde{x}_1+\delta_2, \tilde{x}_2\}$ for each original image. The batch size for AFA was 1024. We decreased the learning rate for AFA by a factor of 0.1 at epochs 300 and 350. The training settings of the vanilla SupCon~\cite{khosla2020supervised} were the same as AFA.

For AdvCL~\cite{fan2021does} and A-InfoNCE~\cite{yu2022adversarial},
we followed the default settings of the original work of A-InfoNCE~\cite{yu2022adversarial}. We adopted the view selection strategy that was the best in the original work. We augmented an image $x$ to $\{\tilde{x}_1, \tilde{x}_2, x+\delta_{cl}, x+\delta_{ce}, HFC(x)\}$ for the multiviewed batch, where $HFC(x)$ is the high-frequency component of $x$. $\delta_{cl}$ is a perturbation that maximizes the contrastive loss. $\delta_{ce}$ is a perturbation that maximizes the cross entropy loss to the cluster of $x$. Those perturbations are applied to the original image. We used batch size = 512 for the contrastive losses.

\smallskip\noindent\textbf{Fine-tuning.}
AFA performs fine-tuning of the linear classifier by the traditional PGD training~\cite{madry2017towards} after pre-training while freezing $\theta_g$, parameters of the feature extractor. We empirically found that adversarial linear fine-tuning shows the best result for our method. We apply adversarial full fine-tuning (AFF) with TRADES $\beta=6$ to AdvCL and A-InfoNCE as AFF was their best-performing fine-tuning method.

For all fine-tuning methods, we used initial learning rate = 0.1, momentum = 0.9, and batch size = 1024. We set the number of attack iterations for the inner maximization to 10. We performed five training epochs on CIFAR10 and 25 epochs on CIFAR100 as a baseline of AFA. We performed 25 epochs of adversarial full fine-tuning for AdvCL and A-InfoNCE, which showed the best performance in the original work. The learning rate of AdvCL and A-InfoNCE decreased by a factor of 0.1 at epochs 15 and 20.

\subsubsection{Traditional Training Methods} \label{sup:settings:1:traditional}

We describe settings to training traditional training methods, including natural training, PGD AT~\cite{madry2017towards}, and TRADES~\cite{zhang2019theoretically}.

\smallskip\noindent\textbf{Adversarial Training.}
We trained a randomly initialized network for 76 epochs for the natural cross-entropy, PGD AT~\cite{madry2017towards}, and TRADES~\cite{zhang2019theoretically}. We used the initial learning rate = 0.1, momentum = 0.9, and weight decay = 1e-4. We decreased the learning rate with a decay factor = 0.1 at epoch 60. The number of attack iterations for the inner maximization was 10. We used the cross entropy loss function for PGD AT and TRADES loss function for TRADES.

\smallskip\noindent\textbf{Natural Training.}
We used the same settings for natural cross-entropy as traditional adversarial training methods except that the training batch and the loss function only consider clean training samples. We used the same setting for the pre-training of vanilla SupCon~\cite{khosla2020supervised} as AFA except that the training batch and the loss function only consider clean training samples. In fine-tuning phase, we applied the standard linear finetuning to the vanilla SupCon.

\subsection{Settings for Section~\ref{sec:evaluation:benchmark}} \label{sup:settings:2}

We used the SGD optimizer for all experiments. We also used weight averaging with decay rate $\tau=0.995$. All experiments were conducted on a machine equipped with eight NVIDIA RTX 4090 GPUs.

\subsubsection{Settings for Wang et al.~\cite{wang2023better}} \label{sup:settings:2:wang}

We followed the training settings of Wang et al.~\cite{wang2023better}. We used 1M samples generated by EDM~\cite{karras2022elucidating} as additional training samples with the original-to-generated ratio of 0.3. That is, 70\% of the samples in a training batch were synthesized and 30\% of them were original. The optimization method was TRADES~\cite{zhang2019theoretically} with $\beta=5$. We used Nesterov momentum for SGD optimization. The initial learning rate, momentum factor, and weight decay factor were set to 0.2, 0.9, and 5 $\times$ 10$^{-4}$, respectively. The learning rate was decreased along with the training epoch by cosine annealing.

\subsubsection{Settings for Adversarial Feature Alignment} \label{sup:settings:2:afa}

We used 1M samples generated by EDM~\cite{karras2022elucidating} and set the original-to-generated ratio to 0.3 for AFA. We set $\lambda_1=1$ and $\lambda_2=5$ for AFA optimization. We did not apply Nesterov momentum to AFA. The initial learning rate, momentum factor, and weight decay factor were set to 0.05, 0.9, and 1 $\times$ 10$^{-4}$, respectively. The learning rate was decreased along with the training epoch by cosine annealing. We used the same augmentation algorithm and augmentation size as Section~\ref{sec:evaluation:baseline} to  Section~\ref{sec:evaluation:ablation}.
\section{Empirical Findings on Alignment} \label{sup:alignment}

\subsection{Separation and Clustering in Input Space} \label{sup:alignment:input_space}

We checked whether the datasets are aligned in the input space. Table~\ref{tab:sup:alignment:input_space} shows the separation and clustering factors measured by distance metrics such as L0, L1, L2, and L-infty. The difference in the scale of the distance metrics causes the difference in the values of the separation factor and clustering factor. Nevertheless, for all distance metrics and all datasets, the train-test clustering factor is much larger than the train-test separation factor, which means that the datasets are not aligned in the input space, i.e., there are samples whose nearest neighbor's class is different from the ground-truth class because the inter-class distance is shorter than the intra-class distance. The distance metric-based 1-nn nearest neighbor classifier also shows low accuracy, illustrating that Yang et al.~\cite{yang2020closer}'s claim that there is no robustness-accuracy tradeoff in the input space does not hold.

\subsection{Separation and Clustering in Feature Space} \label{sup:alignment:feature_space}

We observed the correlation between the feature alignment and the classification accuracy of a neural network in Table~\ref{tab:sup:alignment:feature_space:cifar10} to Table~\ref{tab:sup:alignment:feature_space:res_imagenet}. The maximum clustering factor was larger than the minimum separation factor, indicating that the robustness-accuracy tradeoff is implicit for these feature distributions for all training methods, datasets, and model architectures. The reduction in the accuracy of $f_\text{1-nn}$, i.e., the increase in the number of misaligned samples, in the feature space accompanied the reduction in the accuracy of the neural network regardless of the type of samples.

\section{Additional Experiments} \label{sup:experiments}

\subsection{Empirically Local Lipschitzness} \label{sup:experiments:lipschitz}

Lipschitz constant~\cite{scaman2018lipschitzdnn, fazlyab2019lipschitzacc, chen2020lipschitrelu, huang2021lipschitztrain, kim2021lipschizselfattention} of a linear classifier on the feature space can be used to verify the generalization performance of our method.
We use the empirically local Lipschitz constant $K$ that makes $\frac{|f(x) - f(x')|}{|x - x'|} \leq K$ where $x'=x+\delta$. We perform $l_\infty$ PGD attack to maximize $K$. Table~\ref{tab:evaluation:lipschitz} summarizes the result. Lipschitz bounds of adversarial training methods are smaller than natural training. AFA shows distinctly smaller Lipschitz constants than other methods on CIFAR10, which indicates that the decision boundary of the classifier is smoother for samples within a certain radius by virtue of aligned features. The Lipschitz constant of AFA on CIFAR100 is relatively large, but the accuracy of AFA for worst-case samples was the highest among all methods.

\begin{table}
\begin{footnotesize}
\begin{center}
\caption{Experimental results on empirically local Lipschitzness of a neural network trained by different methods. We report the upper bound $K$ from 10 independent executions such that $\frac{|f(x) - f(x')|}{|x - x'|} \leq K$ for all $x$. Adversarial examples are generated through 20 of PGD iterations following the gradient $\nabla\frac{|f(x) - f(x')|}{|x - x'|}$ with $|\delta|_\infty \leq 8/255$. The columns "Constant" indicate the upper Lipschitz constant $K$, and the columns "Accuracy" indicate the accuracy of the network for $x'$ with the largest $K$.}
\label{tab:evaluation:lipschitz}
\renewcommand{\arraystretch}{1.4}
\begin{tabular}{l||M{1.2cm}M{1.2cm}|M{1.2cm}M{1.2cm}}
\toprule
\textbf{Training}  & \multicolumn{2}{c|}{\textbf{CIFAR10}} & \multicolumn{2}{c}{\textbf{CIFAR100}}\\
\textbf{Method} & \textbf{Constant} & \textbf{Accuracy} & \textbf{Constant} & \textbf{Accuracy}\\ \midrule[0.2ex]
Natural                         &  0.0319 & 46.92 &  0.2687 & 29.56 \\
SupCon                          &  0.2069 & 74.36 &  0.4878 & 48.90 \\
PGD AT~\cite{madry2017towards}  &  0.0089 & 70.17 &  0.0493 & 45.90 \\
AdvCL~\cite{fan2021does}        &  0.0054 & 71.47 &  0.0263 & 49.33 \\
\rowcolor{gray!20}\textbf{AFA (Ours)}   & \textbf{0.0036} & \textbf{76.59} & 0.0623 & \textbf{53.33} \\
\bottomrule
\end{tabular}
\end{center}
\end{footnotesize}
\end{table}

\subsection{Ablation Study} \label{sup:experiments:ablation}

\begin{table}
\begin{footnotesize}
\begin{center}
\caption{Clean and robust accuracies of a neural network trained by AFA with different coefficients $\lambda$. The target dataset and model are CIFAR10 and ResNet-18, respectively. We performed PGD adversarial training on the linear classifier as fine-tuning. The PGD examples were generated with $\epsilon=8/255$, step size $\alpha=2/255$, and attack iteration $N=20$ in test time. The baseline setting of AFA is highlighted in bold.}
\label{tab:evaluation:lambda}
\renewcommand{\arraystretch}{1.4}
\begin{tabular}{l|l||M{1.0cm}M{1.0cm}}
\toprule
\textbf{$\lambda_1$} & \textbf{$\lambda_2$} & \textbf{Clean (\%)} & \textbf{PGD (\%)} \\ \midrule[0.2ex]
0 & 1 & 90.94 & 54.07 \\
1 & 8 & 88.98 & 53.58 \\
1 & 4 & 89.83 & 55.39 \\
1 & 2 & \textbf{91.01} & \textbf{57.77} \\
1 & 1 & 87.19 & 53.48 \\
2 & 1 & 78.81 & 54.28 \\
4 & 1 & 57.52 & 42.66 \\
8 & 1 & 31.60 &  2.37 \\
1 & 0 & 31.51 &     0 \\
\bottomrule
\end{tabular}
\end{center}
\end{footnotesize}
\end{table}

\begin{table}
\begin{footnotesize}
\begin{center}
\caption{Accuracy (\%) and pretraining time of adversarial feature alignment with different contrastive views for ResNet-18 on CIFAR10. We applied baseline training settings except for the contrastive view. The PGD examples were generated with $\epsilon=8/255$, step size $\alpha=2/255$, and attack iteration $N=20$. $T$ is the random image transformation, and $\delta$ is the adversarial perturbation. The baseline contrastive view of AFA is highlighted in bold.}
\label{tab:evaluation:viewsize}
\renewcommand{\arraystretch}{1.4}
\begin{tabular}{l||M{0.9cm}M{0.9cm}|M{1.2cm}}
\toprule
\textbf{Contrastive Views} & \textbf{Clean} & \textbf{PGD} & \textbf{Time (s)} \\ \midrule[0.2ex]
$T_1(x)+\delta$, $T_2(x)$        & 86.55 & 43.64 & 116.58 \\
$T_1(x)+\delta_1$, $T_2(x)+\delta_2$        & 78.91 & 52.55 & 115.82 \\

$x+\delta$, $T_1(x)$, $T_2(x)$        & 64.73 & 38.13 & 191.34 \\
\rowcolor{gray!10}
\textbf{$x+\delta_1$, $T_1(x)+\delta_2$, $T_2(x)$}        & \textbf{88.93} & \textbf{57.68} & \textbf{194.19} \\
\bottomrule
\end{tabular}
\end{center}
\end{footnotesize}
\end{table}

\smallskip\noindent\textbf{Effectiveness of coefficient $\lambda$.} In the optimization problem of AFA, $\lambda_1$ accurately identifies adversarial examples in the feature space and guides them toward alignment. $\lambda_2$ is required for the effective separation and clustering of adversarial examples. We evaluated the accuracy of AFA with different coefficient $\lambda$ values through Table~\ref{tab:evaluation:lambda}. The baseline setting of AFA achieved balanced clean and robust accuracies. On the other hand, increasing the coefficient $\lambda_2$, i.e., $\lambda_1/\lambda_2$ becomes smaller, for adversarial examples resulted in a slight increase in clean accuracy and a slight decrease in robust accuracy. Increasing the coefficient $\lambda_1$, i.e., $\lambda_1/\lambda_2$ becomes greater, for vanilla supervised contrastive learning resulted in a significant decrease in accuracy, showing that the vanilla approach is not compatible with adversarial finetuning. This is because the classes are not sufficiently clustered and separated from each other, resulting in confusing feature values for the linear classifier when performing fine-tuning. On the other hand, AFA's new loss function alone (with $\lambda_1$=0 and $\lambda_2$=1) achieves high accuracy, but combining AFA with vanilla SupCon maximizes performance.

\smallskip\noindent\textbf{Performance of different view sizes.} We identified the impact of the configuration of contrastive views on accuracy in Table~\ref{tab:evaluation:viewsize}. For the same view size, AFA achieved higher robust accuracy when we increased the number of samples perturbed. However, if all samples in a batch are adversarial examples, clean accuracy may suffer. We also see that increasing the view size did not definitely improve accuracy.
While AFA employs more loss terms than standard adversarial training techniques, slowing it down slightly, the view size of AFA is significantly smaller than that of prior adversarial contrastive learning methods such as AdvCL~\cite{fan2021does} and A-InfoNCE~\cite{yu2022adversarial}, leading to a training speed that is more than twice as fast.
Despite the smaller view size, AFA demonstrated superior performance compared to these methods.

\end{document}